\documentclass[pdflatex,sn-basic]{sn-jnl}% Basic Springer Nature Reference Style/Chemistry Reference Style

\usepackage{txfonts}
\usepackage{amsmath}
\usepackage{amssymb}
\usepackage{amsthm}

\jyear{2022}%

\theoremstyle{thmstyleone}%
\theoremstyle{thmstyletwo}%

\theoremstyle{thmstylethree}%

\begin{document}

\title[AI System for ECG Images]{Artificial Intelligence System for Detection and Screening of Cardiac Abnormalities using Electrocardiogram Images}


\author[1]{\fnm{Deyun} \sur{Zhang}}
\author[1,2]{\fnm{Shijia} \sur{Geng}}
\author[3]{\fnm{Yang} \sur{Zhou}}
\author[2,4]{\fnm{Weilun} \sur{Xu}}
\author[1]{\\\fnm{Guodong} \sur{Wei}}
\author[1]{\fnm{Kai} \sur{Wang}}
\author[1]{\fnm{Jie} \sur{Yu}}
\author[1]{\fnm{Qiang} \sur{Zhu}}
\author[1]{\\\fnm{Yongkui} \sur{Li}}
\author[1]{\fnm{Yonghong} \sur{Zhao}}
\author[1]{\fnm{Xingyue} \sur{Chen}}
\author[5,1]{\fnm{Rui} \sur{Zhang}}
\author[1]{\\\fnm{Zhaoji} \sur{Fu}}
\author[1]{\fnm{Rongbo} \sur{Zhou}}
\author[1]{\fnm{Yanqi} \sur{E}}
\author[6,7]{\fnm{Sumei} \sur{Fan}}
\author[8]{\\\fnm{Qinghao} \sur{Zhao}}
\author[5,9]{\fnm{Chuandong} \sur{Cheng}}
\author[9]{\fnm{Nan} \sur{Peng}}
\author[10]{\\\fnm{Liang} \sur{Zhang}}
\author[11]{\fnm{Linlin} \sur{Zheng}}
\author[12]{\fnm{Jianjun} \sur{Chu}}
\author[13]{\fnm{Hongbin} \sur{Xu}}
\author[14]{\\\fnm{Chen} \sur{Tan}}
\author[8]{\fnm{Jian} \sur{Liu}}
\author[15]{\fnm{Huayue} \sur{Tao}}
\author[16]{\fnm{Tong} \sur{Liu}}
\author[16]{\\\fnm{Kangyin} \sur{Chen}}
\author[17]{\fnm{Chenyang} \sur{Jiang}}
\author[3]{\fnm{Xingpeng} \sur{Liu}}
\author*[18,19]{\\\fnm{Shenda} \sur{Hong}}\email{hongshenda@pku.edu.cn}

\affil[1]{\orgname{HeartVoice Medical Technology}, \orgaddress{\city{Hefei}, \country{China}}}
\affil[2]{\orgname{HeartRhythm-HeartVoice Joint Laboratory}, \orgaddress{\city{Beijing}, \country{China}}}
\affil[3]{\orgdiv{Heart Center}, \orgname{Beijing Chao-Yang Hospital, Capital Medical University}, \orgaddress{\city{Beijing}, \country{China}}}
\affil[4]{\orgname{HeartRhythm Medical}, \orgaddress{\city{Beijing}, \country{China}}}
\affil[5]{\orgdiv{Division of Life Sciences and Medicine}, \orgname{University of Science and Technology of China}, \orgaddress{\city{Hefei}, \country{China}}}
\affil[6]{\orgdiv{Department of Physiology}, \orgname{School of Basic Medical Sciences, Anhui Medical University}, \orgaddress{\city{Hefei}, \country{China}}}
\affil[7]{\orgdiv{Department of Anatomy and Histoembryology}, \orgname{School of Basic Medical Sciences, Anhui Medical University}, \orgaddress{\city{Hefei}, \country{China}}}
\affil[8]{\orgdiv{Department of Cardiology}, \orgname{Peking University People’s Hospital}, \orgaddress{\city{Beijing}, \country{China}}}
\affil[9]{\orgdiv{Department of Neurosurgery}, \orgname{The First Affiliated Hospital of University of Science and Technology of China}, \orgaddress{\city{Hefei}, \country{China}}}
\affil[10]{\orgdiv{Heart Center}, \orgname{Anhui Chest Hospital, Anhui Medical University}, \orgaddress{\city{Hefei}, \country{China}}}
\affil[11]{\orgdiv{Department of Electrocardiography}, \orgname{The First Affiliated Hospital of Anhui Medical University}, \orgaddress{\city{Hefei}, \country{China}}}
\affil[12]{\orgname{The Second People’s Hospital of Hefei City}, \orgaddress{\city{Hefei}, \country{China}}}
\affil[13]{\orgdiv{Information Center/Medical Big Data Office}, \orgname{The First Affiliated Hospital of Anhui Medical University}, \orgaddress{\city{Hefei}, \country{China}}}
\affil[14]{\orgdiv{Department of Cardiology}, \orgname{Hebei Yan Da Hospital}, \orgaddress{\city{Langfang}, \country{China}}}
\affil[15]{\orgdiv{Network Information Center}, \orgname{Second Hospital of Tianjin Medical University}, \orgaddress{\city{Tianjin}, \country{China}}}
\affil[16]{\orgdiv{Tianjin Key Laboratory of Ionic-Molecular Function of Cardiovascular Disease, Department of Cardiology, Tianjin Institute of Cardiology}, \orgname{Second Hospital of Tianjin Medical University}, \orgaddress{\city{Tianjin}, \country{China}}}
\affil[17]{\orgdiv{Key Laboratory of Cardiovascular Intervention and Regenerative Medicine of Zhejiang Province, Department of Cardiology}, \orgname{Sir Run Run Shaw Hospital, Zhejiang University School of Medicine}, \orgaddress{\city{Hangzhou}, \country{China}}}
\affil[18]{\orgdiv{National Institute of Health Data Science}, \orgname{Peking University}, \orgaddress{ \city{Beijing}, \country{China}}}
\affil[19]{\orgdiv{Institute of Medical Technology}, \orgname{Health Science Center of Peking University}, \orgaddress{ \city{Beijing}, \country{China}}}

% \affil[2]{\orgdiv{Department}, \orgname{Organization}, \orgaddress{\street{Street}, \city{City}, \postcode{10587}, \state{State}, \country{Country}}}

% \affil[3]{\orgdiv{Department}, \orgname{Organization}, \orgaddress{\street{Street}, \city{City}, \postcode{610101}, \state{State}, \country{Country}}}

\abstract{The artificial intelligence (AI) system has achieved expert-level performance in electrocardiogram (ECG) signal analysis. However, in underdeveloped countries or regions where the healthcare information system is imperfect, only paper ECGs can be provided. Analysis of real-world ECG images (photos or scans of paper ECGs) remains challenging due to complex environments or interference. In this study, we present an AI system developed to detect and screen cardiac abnormalities (CAs) from real-world ECG images. The system was evaluated on a large dataset of 52,357 patients from multiple regions and populations across the world. On the detection task, the AI system obtained area under the receiver operating curve (AUC) of 0.996 (hold-out test), 0.994 (external test 1), 0.984 (external test 2), and 0.979 (external test 3), respectively. Meanwhile, the detection results of AI system showed a strong correlation with the diagnosis of cardiologists (cardiologist 1 (R=0.794, \textit{p}$<$1e-3), cardiologist 2 (R=0.812, \textit{p}$<$1e-3)). On the screening task, the AI system achieved AUCs of 0.894 (hold-out test) and 0.850 (external test). The screening performance of the AI system was better than that of the cardiologists (AI system (0.846) vs. cardiologist 1 (0.520) vs. cardiologist 2 (0.480)). Our study demonstrates the feasibility of an accurate, objective, easy-to-use, fast, and low-cost AI system for CA detection and screening. The system has the potential to be used by healthcare professionals, caregivers, and general users to assess CAs based on real-world ECG images.}

% It is worth noting that the detection and screening results of the AI system can be interpreted by scoring the contribution of pixels in the ECG image.




\maketitle
\section{Introduction}

Cardiovascular disease (CVD) is the leading cause of death worldwide \cite{world2019world, tsao2022heart, timmis2022european, zhao2019epidemiology}. Currently, electrocardiography is one of the simplest and most commonly used CVD detection techniques in clinical practice \cite{yanowitz2012introduction,garcia2014introduction}. Electrocardiogram (ECG) plays an important role in the diagnosis, treatment, and prognosis evaluation of arrhythmia, atrioventricular abnormality, coronary heart disease and other heart diseases. With the development of artificial intelligence (AI), the role of computer-assisted intervention (CAI) is becoming more and more important in the field of ECG analysis \cite{lecun2015deep,he2019practical}. Arrhythmia diagnosis with deep neural networks has reached expert level \cite{hannun2019cardiologist}. In addition, AI-based ECG analysis tasks are gradually diversifying \cite{berkaya2018survey,hong2020opportunities}. Basically, current medical AI applications can be divided into two categories. The first is to use AI to assist in the diagnosis of easily diagnosed diseases such as arrhythmia \cite{hong2020cardiolearn,fu2021artificial}, so as to reduce the workload of doctors \cite{hannun2019cardiologist,liu2018signal}. The second is to perform diagnoses that are difficult for doctors to make \cite{attia2019screening,attia2019artificial}. Such tasks are possible because AI models can recognize patterns and learn useful features from input data that the human eye cannot notice. An example of this is screening for asymptomatic left ventricular dysfunction \cite{attia2019screening}. 

%feasibility of end-to-end deep learning methods that can be used to analyze raw ECG signals is proved based on this research.

\begin{figure}[htp]
    \centering
    \includegraphics[width=10cm]{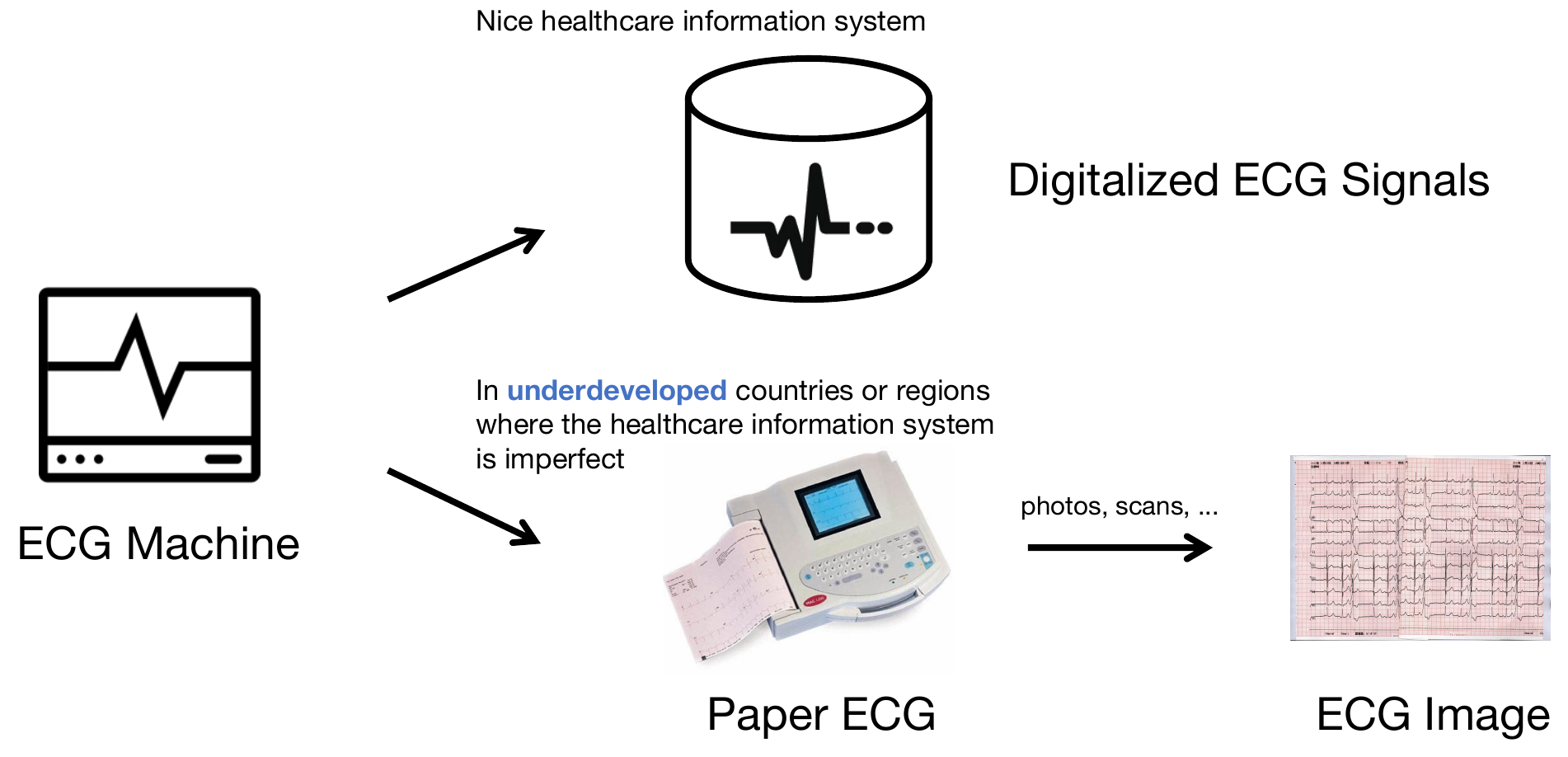}
    \caption{In underdeveloped countries or regions where the healthcare information system is imperfect, it is difficult to obtain ECG data in digital signal format. Patients and doctors only have access to ECG images.}
    \label{fig0}
\end{figure}

However, in \textbf{underdeveloped countries or regions} where the healthcare information system is imperfect, it is difficult to obtain ECG signals in digital signal format. In many cases, patients and doctors only have access to paper ECGs \ref{fig0}. Therefore, analysis methods based on ECG signals are difficult to be applied in these situations. 

ECG images have received attention from researchers over the past few years, and were used for the task of developing image analysis methods. The related studies can be divided into two categories: some researchers digitized ECG images and re-analyzed them using ECG signal analysis methods \cite{li2020deep,baydoun2019high}; other researchers directly extracted image features from ECG images for analysis \cite{sangha2022automated,yu2018qrs,irmak2022covid}. However, both approaches are difficult to build into mobile systems that can be used in the real world. The main challenges are as follows: 
\begin{itemize}
\item ECG images have higher data dimensions than ECG signals. Due to the high-dimensional nature, mobile devices cannot meet the computing resource requirements of AI algorithms for analyzing ECG images.
\item There is a gap between the carefully constructed datasets of most previous studies and ECG images obtained directly from the real world. Reasons for the discrepancies include camera angle, pixel distortion, presence of other irrelevant objects in the image, etc. 
\item The real world contains a large amount of ECG images. However, the data available for direct analysis is very limited. Due to these challenges, comprehensive evaluations of AI-based mobile systems to analyze and process ECG images directly in the real world are lacking.
\end{itemize}
 
It is worth noting that the COVID-19 pandemic has led to another rise in the incidence of CAs \cite{nchioua2022strong, stein2022sars, yuk2022heart, zhang2022data}. Therefore, countries or regions with insufficient medical resources or underdeveloped medical information systems urgently need an accurate, objective, easy-to-use, fast, and low-cost CVD detection and screening tool.

\begin{figure}[htp]
    \centering
    \includegraphics[width=11cm]{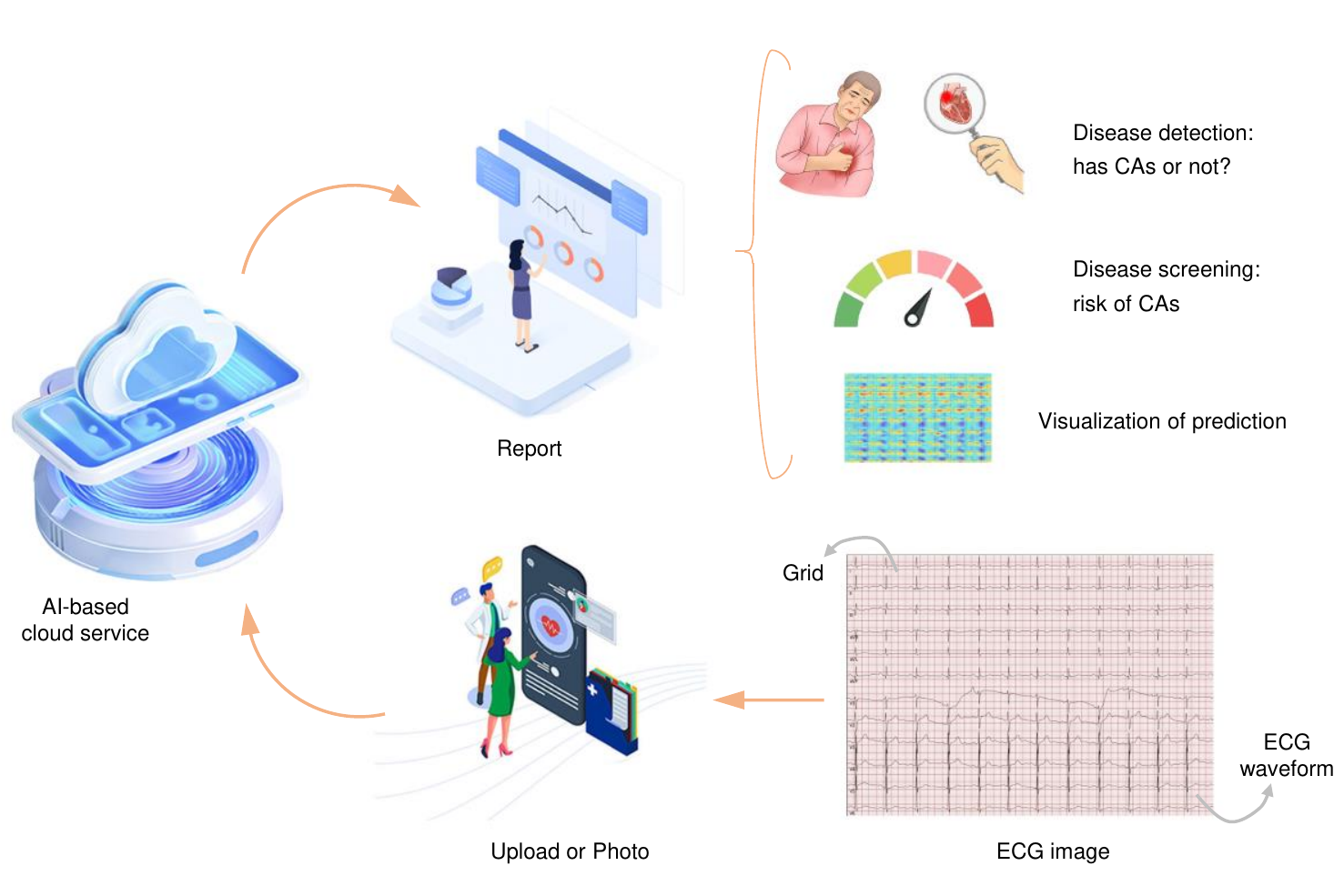}
    \caption{Overview of the AI system for CA detection and screening from real-world ECG images. The system uses the images uploaded or taken by the user to obtain the ECG waveform region to be analyzed. An AI-based cloud service is used to detect the presence of CAs in submitted ECG images. In addition, the risk of CAs can also be provided based on the ECG waveform region. It is worth noting that the contribution score of each pixel in the obtained ECG waveform region can be visualized according to the analysis results.}
    \label{fig1}
\end{figure}

In our study, we developed an AI system for the detection and screening of CAs using ECG images, which is the first of its kind to our knowledge (Fig \ref{fig1}, Extended data Fig. \ref{figext1}, Extended data Fig. \ref{figext2}). The system receives ECG images, which can be collected using photos or scans of paper-based ECGs. An important component of the system is the section that automatically displays gradient-weighted class activation maps based on detection and screening results, which helps in interpreting the model outputs. Our system is designed to provide an accurate, objective, non-invasive, simple, fast, and low-cost method for the diagnosis and screening of CAs. Users can obtain analysis results through mobile phones with Internet access all over the world. 

\section{Results}

\begin{figure}[htp]
    \centering
    \includegraphics[width=9cm]{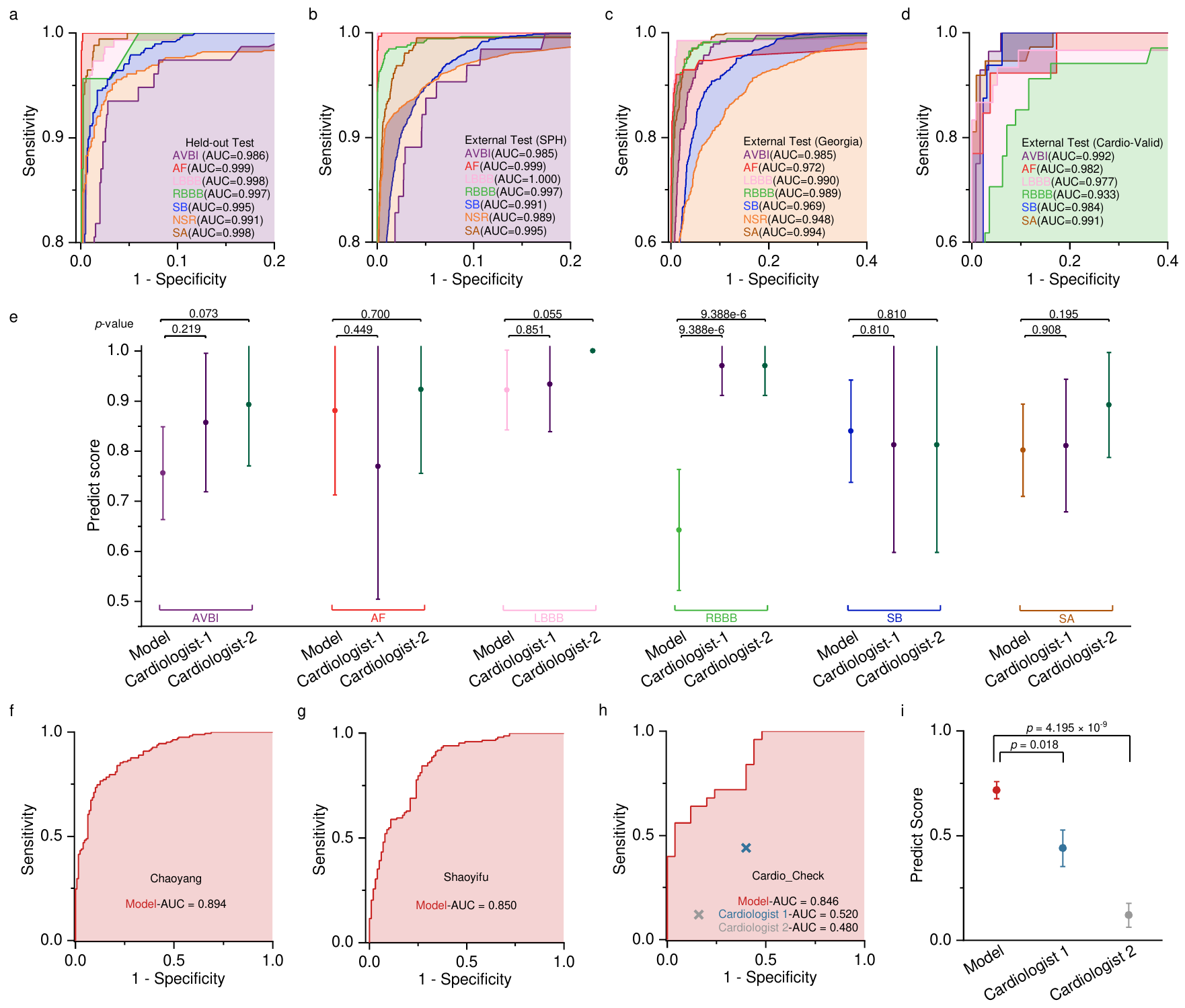}
    \caption{{\bfseries{Performance of CA detection and screening from ECG images.}} {\bfseries{a}}, ROC curves for detecting CAs from ECG images in the hold-out test dataset (n = 1640 ECG images from 1535 patients). {\bfseries{b}}, ROC curves for detecting CAs from ECG images in the SPH dataset (n = 16567 ECG images from 16460 patients). {\bfseries{c}}, ROC curves for detecting CAs from ECG images in the Georgia dataset (n = 2871 ECG images from 2871 patients). {\bfseries{d}}, ROC curves for detecting CAs from ECG images in the Cardio-valid dataset (n = 146 ECG images from 146 patients). {\bfseries{e}}, Interval plots of AI model and cardiologist on the CA detection task. The point in the middle of each interval represents the mean, and the upper and lower edges of the interval range represent the 95\%CI. {\bfseries{f}}, CA screening performance on the Chaoyang dataset (n = 290 ECG images from 112 patients). {\bfseries{g}}, CA screening performance on the Shaoyifu dataset (n = 248 ECG images from 248 patients). {\bfseries{h}}, Performance comparison of AI model and cardiologist (n = 50 ECG images from 46 patients). {\bfseries{i}}, Interval plots of AI model and cardiologist in CA screening task. AVBI, first-degree atrioventricular block. AF, atrial fibrillation. LBBB, left bundle branch block. RBBB, right bundle branch block. SB, sinus bradycardia. NSR, sinus rhythm. SA, sinus tachycardia. The point in the middle of each interval represents the mean, and the upper and lower edges of the interval range represent the 95\%CI. In our analysis, our model demonstrated robust performance on data from diverse institutions and populations. Compared with cardiologists, the predictions of our model did not differ significantly from the diagnoses of cardiologists and showed strong agreement (Extended data Fig. \ref{figext9},\ref{figext10}). Our model still shows competitive results for screening task that is difficult for cardiologists to achieve. The \textit{p} values are displayed above the boxplots.}
    \label{fig2}
\end{figure}

{\bfseries{Evaluation of detection and screening for CAs in ECG images}} We evaluated the accuracy of CA diagnosis and AF screening based on a large number of ECG images. The receiver operating characteristic (ROC) curves for CAs diagnostic and AF screening tasks are shown in Fig. \ref{fig2}a,f. For CA detection tasks, the weighted mean AUC, sensitivity (SEN), specificity (SPE), and F1 score (F1) of our model were 0.996 (95\% confidence interval (CI), 0.993-0.999), 0.889 (95\% CI, 0.875-0.904), 0.995 (95\% CI, 0.992-0.998) and 0.902 (95\% CI, 0.8888-0.916), respectively (Table \ref{tab1}, Extended data Table \ref{extab2}). Extended data Fig. \ref{figext3} further demonstrated the independent ROC curve, decision curve analysis (DCA) , and confusion matrix of our model for different types of diseases in the detection task. For screening tasks, the weighted mean AUC, SEN, SPE, and F1 of our model were 0.894 (95\% CI, 0.859-0.929), 0.833 (95\% CI, 0.790-0.876), 0.789 (95\% CI, 0.742-0.836) and 0.833 (95\% CI, 790-0.876), respectively (Table \ref{tab1}, Extended data Table \ref{extab2}). The DCA for screening task is shown in Extended data Fig. \ref{figext5}a. The confusion matrix for screening task is shown in Extended data Fig. \ref{figext5}b. The boxplots and normal distribution curves for screening task are shown in Extended data Fig. \ref{figext5}c. We found that the model had a different distribution of predictions for different AF risk populations.

\begin{table}[htp]
	\caption{AUCs of the AI Model on different datasets. Numbers in parentheses represent 95$\%$CI.}
	\centering
	\scalebox{0.6}{
	%% \tablesize{} %% You can specify the fontsize here, e.g., \tablesize{\footnotesize}. If commented out \small will be used.
	\begin{tabular}{cccccccc}\\
	\hline
	Task & Label & Hold-out test & SPH & Georgia & Cardio-valid & Chaoyang & Shaoyifu\\
	\hline
	CA detection & AVBI & 0.986(0.981-0.992) & 0.985(0.983-0.987) & 0.985(0.980-0.989) & 0.992(0.979-1.000) & ------ & ------\\
	& AF                & 1.000(0.999-1.000) & 1.000(1.000-1.000) & 0.972(0.966-0.978) & 0.982(0.961-1.000) & ------ & ------\\
	& LBBB              & 0.998(0.995-1.000) & 1.000(1.000-1.000) & 0.990(0.987-0.994) & 0.977(0.954-1.000) & ------ & ------\\
	& RBBB              & 0.997(0.994-1.000) & 0.997(0.996-0.998) & 0.989(0.985-0.993) & 0.933(0.894-0.972) & ------ & ------\\
	& SB                & 0.995(0.992-0.999) & 0.991(0.989-0.992) & 0.969(0.963-0.975) & 0.984(0.965-1.000) & ------ & ------\\
	& NSR               & 0.991(0.987-0.996) & 0.989(0.988-0.991) & 0.948(0.940-0.956) & ------ & ------ & ------\\
	& SA                & 0.998(0.997-1.000) & 0.995(0.994-0.996) & 0.994(0.991-0.996) & 0.991(0.976-1.000) & ------ & ------\\
	& Weighted mean & 0.996(0.993-0.999) & 0.994(0.992-0.995) & 0.984(0.980-0.989) & 0.902(0.888-0.916) & ------\\
	\hline
	CA screening & YAF & ------ & ------ & ------ & ------ & 0.894(0.859-0.929) & 0.850(0.806-0.894) \\
	\hline
	\end{tabular}}\label{tab1}
\end{table}

We further tested whether integrating the results of multiple ECG images from the same patient could improve the accuracy of AF screening. The Chaoyang dataset was used in this study because it contains multiple ECG images from a single patient (mean(SD), 3(2)). The risk of AF on ECG images was calculated and integrated for all patients (Extended data Fig. \ref{figext4}). In this data, the sensitivity and specificity of patients with AF and patients without AF were further increased from 0.833 and 0.789 to 0.876 and 0.833, respectively (Extended data Fig. \ref{figext5}i). In addition, the integrated ROC increased from 0.894 to 0.929 (Extended data Fig. \ref{figext5}g,i). The clinical benefit was improved compared with before the integration (Extended data Fig. \ref{figext5}h). The results showed that multiple screening could improve the accuracy of CA screening results.

{\bfseries{Generalization in the external test cohort}} To evaluate the generality of our model across different institutions and patient populations, we validated our model on multiple external test cohorts. The CA detection model was validated in SPH (Asian) datasets and Georgia (Occidental) datasets. Our models obtained 0.994 (95\%CI, 0.992-0.995, SPH) and 0.984 (95\%CI, 0.980-0.989, Georgia) AUCs, respectively (Table \ref{tab1}, Extended Data Table \ref{extab2}). The performance indicates that our model could be extended to CA detection tasks in different populations around the world (Extended data Fig. \ref{figext6}, Extended data Fig. \ref{figext7}). 

The CA screening model was validated in Shaoyifu datasets. Unlike the Chaoyang dataset, which includes data from people in northern China, the Shaoyifu dataset includes ECG images from people in eastern China. Our models obtained 0.850 (0.806-0.894) AUC (Table \ref{tab1}, Fig. \ref{fig2}g). The DCA is shown in Extended Data Fig. \ref{figext5}d. The confusion matrix for is shown in  Extended Data Fig. \ref{figext5}e. The boxplots and normal distribution curves for screening task is shown in Extended data Fig. \ref{figext5}c. The results show that our model can still accurately distinguish data it has never seen during training.

{\bfseries{Performance comparison with cardiologists}} Based on external tests, we further compared the performance difference between our model and cardiologists. We measured the performance of our model in detecting CA using 146 ECG data from the TNMG network from April to September 2018 \cite{sangha2022automated} and compared it with cardiologists. We found that there was no significant statistical difference between the model and the cardiologist except for the prediction results of RBBB (\textit{p}=9.388e-6) (Fig. \ref{fig2}e). This difference might be due to disease subtypes (e.g., incomplete right bundle branch block (IRBBB) vs. complete right bundle branch block (CRBBB)). Overall, the performance showed that our model was close to or even on par with cardiologists in detecting CA on ECG images (Extended Data Fig. \ref{figext8},\ref{figext9},\ref{figext10}).

Unlike the detection of CAs, there are currently no publicly available data to assess a patient's risk for AF. Therefore, 50 sinus ECG images from the Chaoyang dataset (25 from AF patients and 25 without AF ) were randomly selected and provided in double-blind to two cardiologists for screening. We found that identifying the risk of AF from sinus rhythm ECG was a difficult task for experienced cardiologists (Fig. \ref{fig2}h,i). Our model had a higher AUC than cardiologists (0.846(model) vs 0.52(cardiologist 1) vs 0.48(cardiologist 2)). Our model could provide an objective, non-invasive, simple, quick, and low-cost way to screen for AF. Meanwhile, the screening method could easily be extended to screening for other CAs.

{\bfseries{Interpretability of AI model}} External test results showed that our model could still perform well on data that has never been seen. However, the interpretability of our model needed to be provided to further support its detection/screening plausibility. In our model, we introduced the gradient-weighted class activation map (Grad-CAM) \cite{selvaraju2017grad} to score the contribution of each pixel in the ECG image according to the predicted results. Since the ECG image contains all the information needed for analysis, we could provide interpretability for the model by analyzing the contribution score of the pixels in the ECG image. At the same time, the pixel contribution score could also assess whether the key information affecting the prediction results of our model is based on clinically relevant information or spurious data features \cite{degrave2021ai}.

\begin{figure}[htp]
    \centering
    \includegraphics[width=11cm]{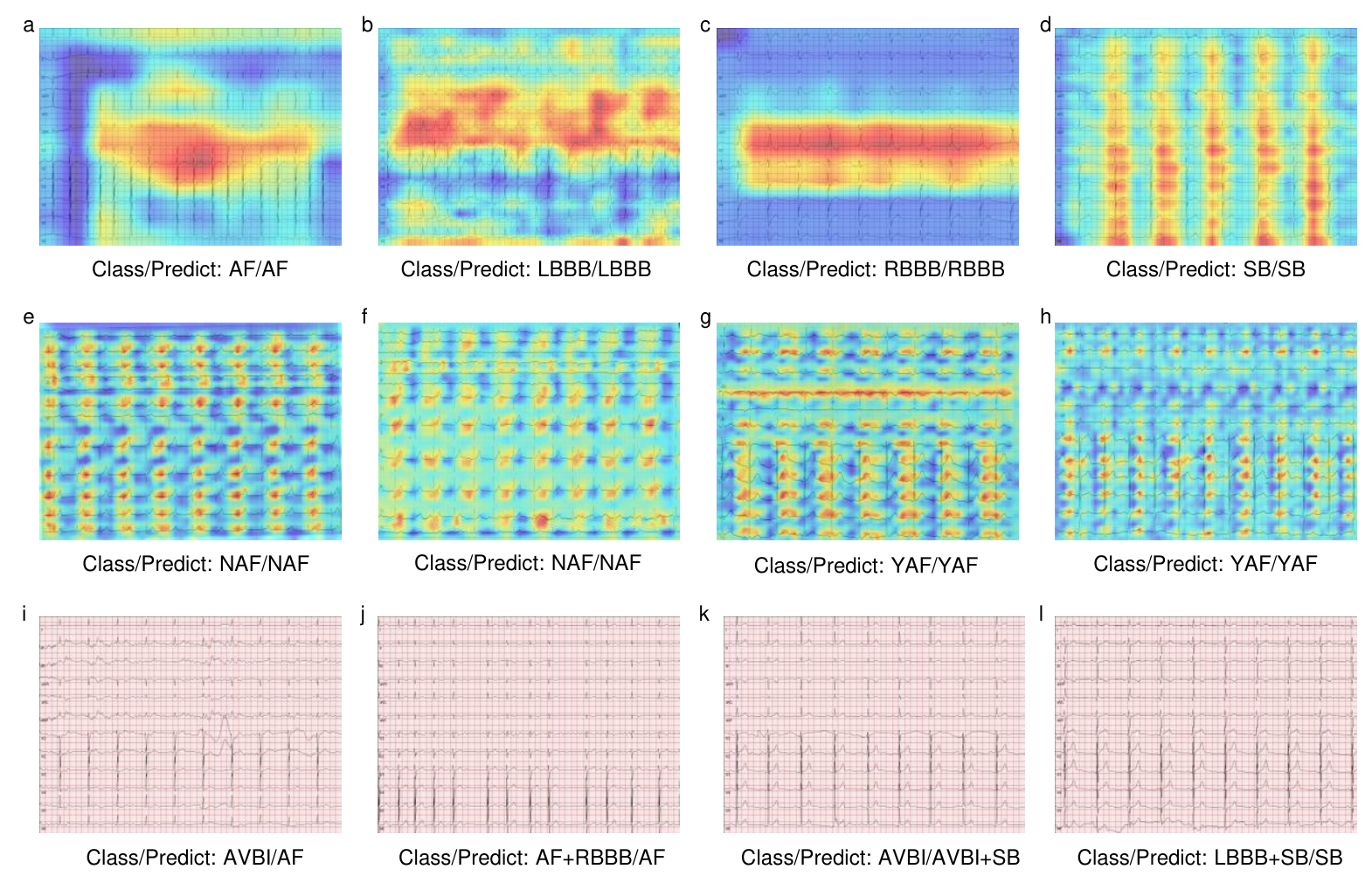}
    \caption{{\bfseries{Grad-CAM visualization results and mispredicted samples of our model.}} {\bfseries{a-h}}, Examples of Grad-CAM saliency maps superimposed on ECG images are shown. The ground truth and prediction result of the ECG images are displayed below the ECG images. In the CA detection task, the visual consistency between the judgment basis of our model and the corresponding ECG changes was presented. In the CA screening task, our model paid attention on some subtle changes in sinus ECG images. {\bfseries{i-l}} , Samples of prediction error from our model. Like cardiologists, our model predictions were inaccurate due to noise interference and waveform changes.}
    \label{fig3}
\end{figure}

We found that the model had different attention regions for different ECG images. Irregular F-wave replacement of P-wave in lead V1 is most evident during AF episodes (Fig. \ref{fig3}a). There were significant changes in QRS, LBBB and RBBB compared with normal sinus rhythm (Fig. \ref{fig3}b,c). The slow heart rate of SB allowed the model to focus more on the RR interval (Fig. \ref{fig3}d). Interestingly, the areas of concern in our model for the detection of CAs had extremely high consistency with the corresponding waveform changes in CAs. These findings mean that our model could detect changes in waveform from ECG images. In addition, the pixel importance visualization method can assist doctors in diagnosis.

Screening for AF has long been one of the challenges facing the medical field. Therefore, we further explored the ECG images of sinus rhythm in patients with and without AF. We found that the model paid the most attention to the T-wave region, followed by the P-wave (Fig. \ref{fig3}e,f,g,h). This was consistent with the previous research \cite{baek2021new}. Studies had shown that the key point to detect subtle changes in paroxysmal AF is within 0.24s of QRS wavefront of 12-lead ECG. In addition, the region of concern in our model did not correspond to specific leads and appeared to be shown in all leads. This could provide a key evidence for AF screening. In the future, the AI predicted risk value of AF in sinus rhythm is expected to become an important biomarker.

{\bfseries{Case study of misprediction}} Although the model could realize the detection of CAs in ECG images, some errors will inevitably occur. Therefore, we further studied the samples that detected the error. Similar to the signal analysis model, our model also produced false predictions when subjected to strong noise interference (Fig. \ref{fig3}i). In addition, dramatic changes in the waveform or interval caused by the onset of arrhythmia can increase the difficulty in detecting CA (Fig. \ref{fig3}j,k). Similar to human experts, our model would also produce an incorrect prediction when the waveform changes are not obvious (Fig. \ref{fig3}l).

{\bfseries{UMAP visualization}} Based on the visualization of the Grad-CAM, we further studied the effectiveness of the features extracted by the model. We used UMAP to convert the high-dimensional feature vector in the AI model into two dimensions. After that, two-dimensional arrays were plotted as scatter diagrams. Finally, the scatter diagram was colored according to the category to realize visualization.

\begin{figure}[htp]
    \centering
    \includegraphics[width=11cm]{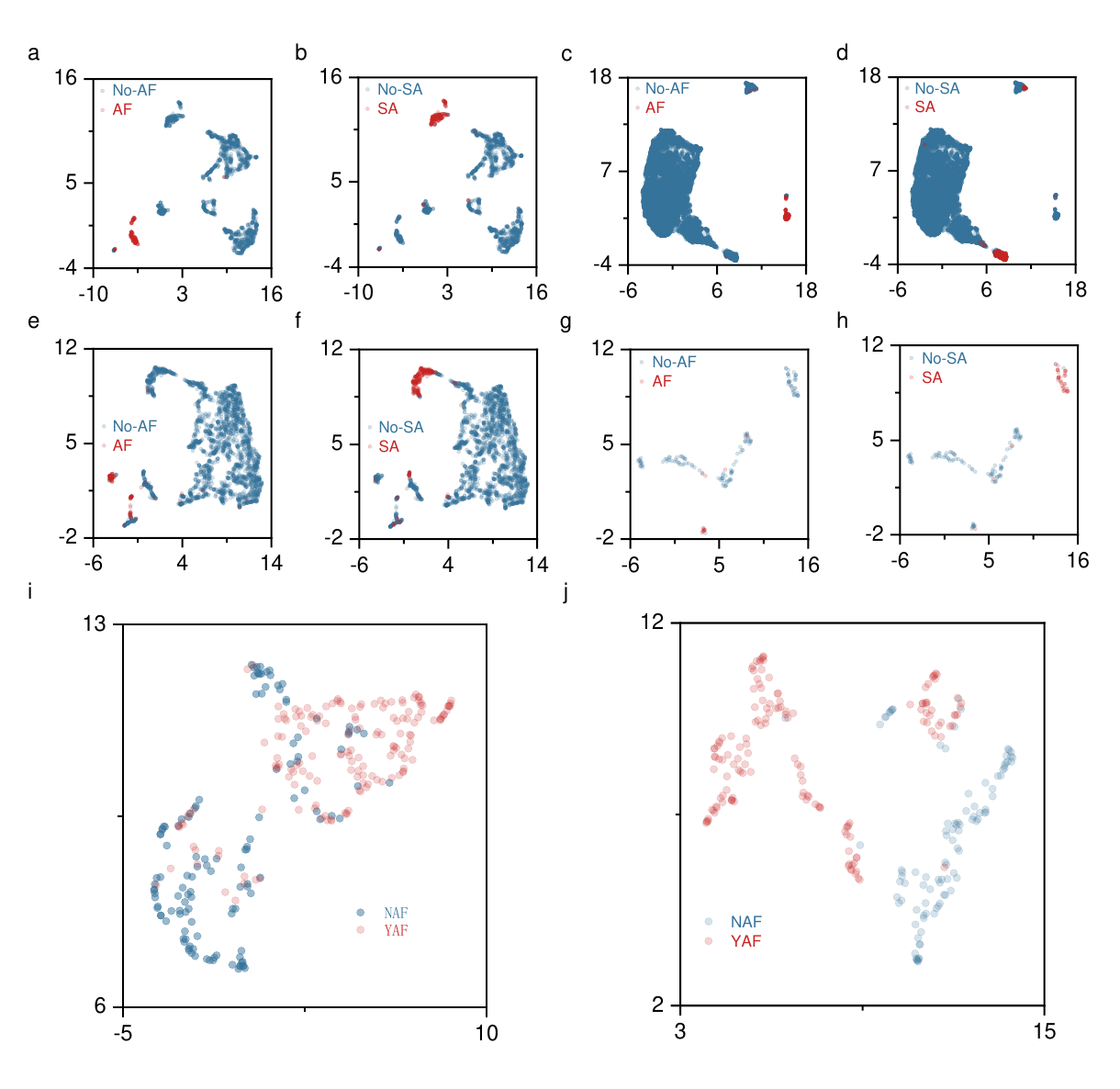}
    \caption{{\bfseries{Visual scatter plot of features extracted by our model after UMAP dimension reduction.}} {\bfseries{a-h}}, Distributions of AF and SA after dimension reduction on hold-out test dataset, SPH dataset, Georgia dataset, and Cardio-valid dataset are shown, respectively. {\bfseries{i,j}}, In the screening task, low-dimensional data distributions from NAF and YAF after UMAP dimension reduction from Chaoyang and Shaoyifu dataset. The different types of diseases are significantly divided based on the low-dimensional data of UMAP reduction in the scatter plot. The feature distributions show that our model can extract key ECG features from sinus rhythm ECG images.}
    \label{fig4}
\end{figure}

In the task of CA detection, features extracted by our model from different types of ECG images are significantly different (Fig. \ref{fig4}a,b, Extended Data Fig. \ref{figext11}a,b,c,d,e,f,g). Compared with the internal test dataset, the features obtained from the external test dataset had similar differences (Fig. \ref{fig4}c,d,e,f,g,h, Extended Data Fig. \ref{figext11}h,i,j,k,l,m,n, Extended Data Fig. \ref{figext12}). It further proved that the model had good generalization ability. Unlike the CA detection task, screening for AF from sinus rhythm was an extremely difficult task. However, our model can still roughly distinguish AF patients from non-AF patients (Fig. \ref{fig4}i,j).

Although the features extracted by our model have obvious differences, there would still be outliers. We visualized these outliers to evaluate our model more accurately (Extended Data Fig. \ref{figext13}). The features extracted by our model may be biased towards a single disease when multiple diseases are present in the ECG image. This bias would inevitably affect the detection performance of the model for many kinds of diseases and lead to the occurrence of wrong predictions.

{\bfseries{Ablation studies of multi-stage fine-tuning}} To improve the accuracy of screening results in the AF screening task, a two-stage fine-tuning strategy was designed for the training phase of our model (Extended Data  Fig. \ref{figext14}a,b,c). Different datasets were used to train our models at different stages of fine-tuning. The purpose of two-stage fine-tuning was to further strengthen the learning of semantic features in ECG images to adapt to task objectives.

We conducted 20 independent tests at each stage to more comprehensively compare the performance of different models. We performed a statistical analysis based on an independent two-sample t-test for F0, F1, and F2 through 20 independent tests (Extended Data \ref{figext14}d,e,f,g). The results showed that compared with F0, F1 and F2 showed significant differences in each indicator. Among them, F1 and F2 showed no statistical difference, and F1 slightly outperformed F2.

{\bfseries{Performance comparison of model and system on different quality real-world ECG images}} The overall diagnostic flow of the proposed AI system was demonstrated in the Extended Data Fig. \ref{figext2}. All interfaces were as simple as possible by removing unnecessary content, aiming to provide clear and understandable information for the average user (Extended Data Fig. \ref{figext1}). These designs make it easy for users to use the system. Two questions were hard to ignore: (1) Are there performance differences between applications and models? (2) Can ECG images from mobile devices be used for the analysis of applications? To compare the performance differences between the model and system, we constructed real-world ECG image dataset of different qualities and further verified the system on this dataset (implementation details are provided in the Methods section). 

\begin{figure}[htp]
    \centering
    \includegraphics[width=9cm]{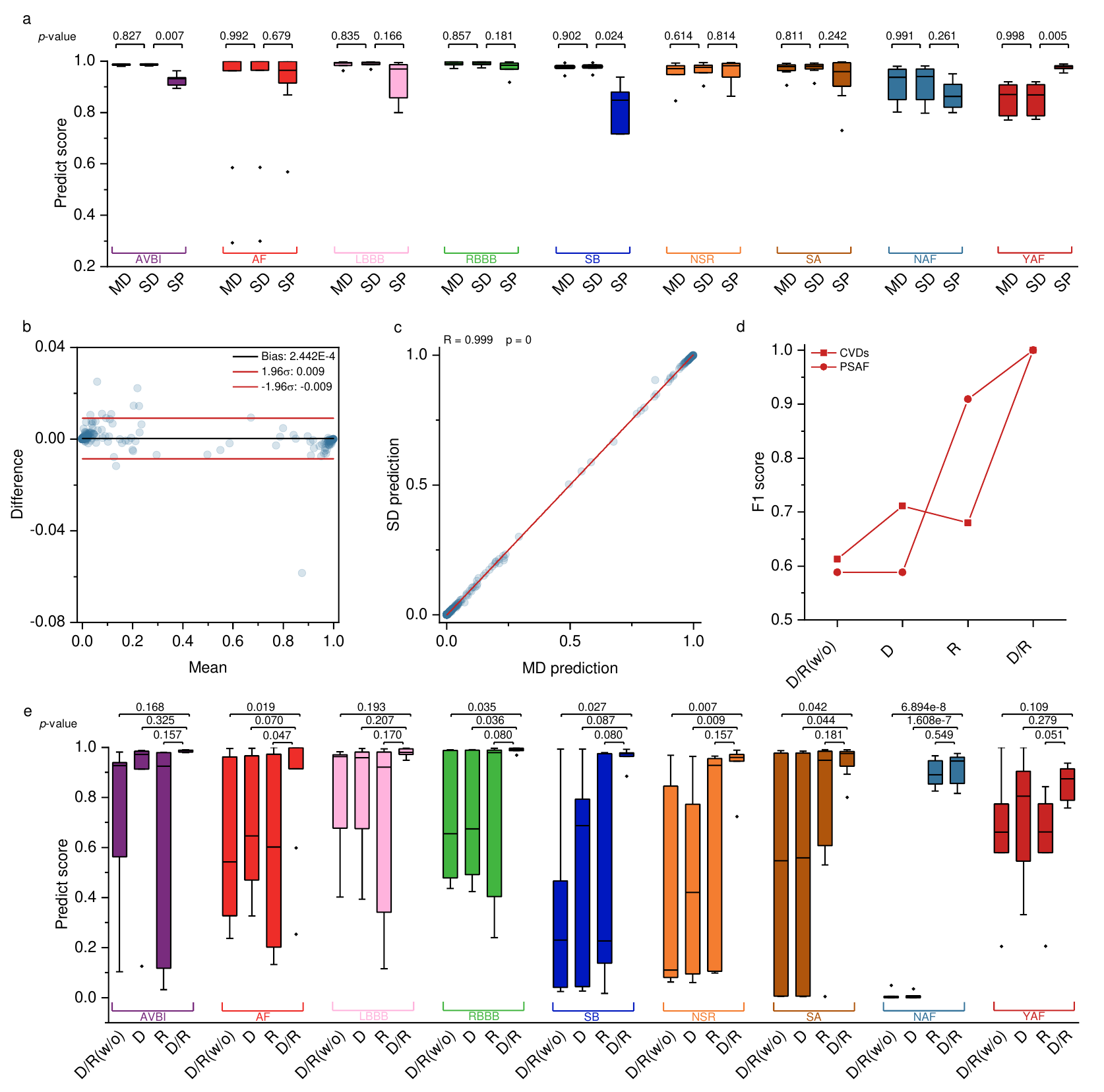}
    \caption{{\bfseries{Ablation experiment results for models, applications, and decision processes.}} {\bfseries{a}}, boxplots of model and application predicts for different qualities of data (digital and photo). MD, the prediction results of our model are based on digital data. SD, the prediction results of our system are based on digital data. SP, the prediction results of our system are based on photo data. {\bfseries{b}}, Bland-Altman plot based on AI model (MD) and AI system (SD). {\bfseries{c}}, Correlation between AI model (MD) and AI system (SD). {\bfseries{d}}, Performance curves with different processing methods. D/R(w/o), CA detection and screening were performed directly on the input images without processing. D, CA detection and screening were performed on the images after waveform region detection. R, CA detection and screening were performed on the images after angle-corrected. D/R, CA detection and screening were performed on the images after waveform region detection and angle-corrected. {\bfseries{e}}, boxplots of predicted results of different data processing methods. AVBI, first-degree atrioventricular block. AF, atrial fibrillation. LBBB, left bundle branch block. RBBB, right bundle branch block. SB, sinus bradycardia. NSR, sinus rhythm. SA, sinus tachycardia. NAF, sinus rhythm without AF. YAF, sinus rhythm before AF. Different diseases represented by different colors are shown below the boxplots in a,e. On each box, the center line represents the median, and the bottom and top edges of the box represent the 25th and 75th percentiles, respectively. The extension beyond the box represents 1.5 times the interquartile range. Outliers are represented using diamonds. The \textit{p} values are shown above the boxplots.}
    \label{fig5}
\end{figure}

High-quality ECG data (digital) were used to measure differences between models and applications (Extended Data Fig \ref{figext15}a,b,c). The comparison results showed that there was no significant difference in performance between model and system (R=0.999, \textit{p}=0) when the quality of the input ECG image was the same (Fig. \ref{fig5}a,b,c). Further, we used iPhone 11 to photograph the printed paper-based ECG to obtain a low-quality ECG image dataset (photo, Extended Data Fig. \ref{figext15}d,e,f,g). We found that the change in image quality would reduce the prediction probability by inputting high-quality data and low-quality data into the system (Fig. \ref{fig5}a). However, data obtained from mobile devices did not result in performance degradation (Fig. \ref{fig5}b,c). Strong consistency (R=0.974, \textit{p}=5.618e-219) was still maintained between high-quality data and low-quality data (Extended Data Fig. \ref{figext15-16}a,b). Real-world validation results demonstrate that the AI analysis method could be used to analyze ECG images directly in the real world and evaluated the feasibility of the AI analysis method for ECG images in healthcare applications.

{\bfseries{Effects of changing imaging conditions in the real world on ECG image analysis results}} We found that the change in image quality would affect the results of ECG image analysis. Lighting, contrast, angle, etc. could have a major impact on image quality when using mobile devices to capture images. Therefore, it was necessary to further evaluate the influence of imaging conditions on ECG analysis results. ECG images obtained under different imaging conditions are shown in the Extended Data Fig. \ref{figext16}a,b,c,d,e,f,g,h. The results show that the effect of rotation angle, ECG ratio, and tilt angle on model performance was less than that of brightness and contrast (Extended Data Fig. \ref{figext17}). 

{\bfseries{Rationality analysis of image processing process}} The cross study of AI techniques could combine the advantages of different methods to reduce the difficulty of data analysis and processing in downstream tasks. Therefore, a series of preprocessing operations (more details are provided in the Methods section) were designed to reduce the difficulty of detecting and screening for CA from ECG images (Fig. \ref{fig5}d,e).

We evaluated the effectiveness of image processing procedure by choosing to add or not add preprocessing operations. We found that system performance decreased when the relevant preprocessing operations were removed (Fig. \ref{fig5}.e). F1 of CA detection and screening tasks increased from about 0.6 to 1.0 with image processing (Fig. \ref{fig5}.d). The results show that preprocessing was very important for ECG image analysis in the real world and can further improve the accuracy of detection and screening(Extended Data Fig. \ref{figext15-16}c-h).

\section{Discussion}

Our study shows that AI can detect CAs from real-world ECG images and further screen the risk of CAs. This is the first comprehensive assessment analyzing ECG images acquired in the real world. Importantly, the results demonstrate the potential of AI to provide an entirely new approach to CA diagnosis and screening. The system not only assists doctors in diagnosis, but also helps them complete difficult tasks (Fig. \ref{fig2},\ref{fig3}). The system is objective and not subject to patient or clinician influence. At the same time, the system provides a non-invasive, simple, and fast detection method. Additionally, the app can be accessed from a mobile device anywhere there is an an Internet connection.

Our research has the potential to reduce the extra time or money patients spend on the treatment. In areas with uneven distribution of medical resources, most high-quality medical resources are concentrated in a few areas \cite{roth2020global}, but patients are distributed in various regions \cite{zhao2019epidemiology}. Therefore, patients in most areas spend a lot of time and energy in the treatment process. In addition, due to the imperfection of medical informatization, patients can only obtain paper-based ECG after ECG examination. If patients want a more accurate diagnosis, they need to visit an outpatient clinic. This approach often imposes unnecessary costs on patients. Our system can reduce the need for outpatient visits, reduce unnecessary costs to patients, and provide expert-level diagnostic services to patients in under-resourced areas by providing a simple, fast, and low-cost way to assess patients' CAs.

Screening for CA is receiving increasing attention, especially for AF \cite{freedman2017screening}. Compared with other medical screening tests, such as cervical cancer Pap smear (AUC 0.70) \cite{chen2016pax1}, the identification of undetected AF (based on ECGs recorded during a normal sinus rhythm) by inexpensive, widely available point-of-care tests is of greater practical interest. Our system can be easily used at the set point of care, and moreover, can be easily generalized to other CA screening tasks.

Medical AI is often criticized for lack of detailed reporting and limited interpretation and reproducibility of results \cite{maier2018rankings}. Our system provides a solution to score the contribution of each pixel in the ECG image to reveal the leads or regions that contribute more to the predicted outcome (Fig. \ref{fig3}a-h). It can assist doctors in diagnosis or perform self-diagnosis. At present, the early identification, diagnosis and timely treatment of AF in sinus rhythm is one of the major challenges facing the medical community. By scoring the contribution of pixels, we found that unlike the high attention of leads, our model pays more attention to T-wave and P-wave regions than other regions. The results show that the key point to detect subtle changes in paroxysmal AF is within 0.24s of the QRS wavefront of the 12-lead ECG \cite{baek2021new}. In addition, Bayes syndrome is associated with the risk of AF and stroke \cite{arboix2017bayes}. These studies are consistent with our model focusing on the P-wave region. Our model focuses surprisingly on the T-wave region. The association between T-wave and AF has not been revealed in previous studies. This is an exciting discovery. In our visualization results, we find that deformations near the T-wave region in ECG images are associated with a higher risk of AF. Therefore, we hypothesize that there may be some subtle changes in the T-wave region in ECG images that cannot be detected by the human eye. By extracting hidden features in images, our model can detect these changes in more detail, and enable the prediction of AF risk from the sinus rhythm. 

The reason why real-world images are difficult to analyze compared to carefully constructed datasets is that images captured in the real world often contain a lot of irrelevant information or noise, such as effective information ratio, angle, brightness, etc. The cross study of AI methods can combine the advantages of different methods and reduce the difficulty of data analysis and processing in downstream tasks \cite{su2020multimodal}. In our study, we introduce methods such as lightweight neural network structure, image recognition, object detection to overcome difficulties and challenges such as differences in data dimensions, interference from natural environments, and too few trainable samples (Extended Data Fig \ref{figext18},\ref{figext19},\ref{figext20}). Overall, with a data-driven analytic strategy, as well as combining multiple AI approaches, our system can remarkably improve the accuracy of CA detection and screening (Fig. \ref{fig5}).

We envision that the system can be used in any setting. Clinicians can use the system to assist in the diagnosis of CA. In addition, public health departments or physical examination institutions can use the system to screen for CA. Individual users can also use the system to conduct CA assessment and consult doctors online. Our future research is to determine the feasibility of this approach and its potential impact on clinical practice.

A possible technical limitation of our study is that the ECG image input was not carefully processed. Unpredictable results can be obtained when the shooting environment is poor or the paper-based ECG image style is changed. In future research, we will add processing to paper-based ECG images, such as removing background grids, binarization, etc. Another limitation of the work is that it is only applicable to the 12X1 layout of ECG images. The performance of our model on 2X6 and 3X4 ECG images needs further improvement (Extended Data Fig \ref{figext21}). Future studies with other ECG image formats are needed to further confirm these results. Furthermore, our study demonstrates that our system can also screen for AF. Unfortunately, due to data limitations, we did not evaluate the performance of our model on other CA screening tasks. Finally, we found a certain correlation between the onset of AF and P waves, as well as T waves, which is consistent with previous studies. However, the specific mechanism causing the onset of AF has not been fully elucidated and further studies are needed.

In conclusion, our work demonstrates that AI can provide a new system for CA diagnosis and screening, and allow the development of new biomarkers to support CA screening. The system can be applied globally without imposing additional burdens on users. It is of great significance to the diagnosis, treatment, and management of CA patients.

\section{Methods}

{\bfseries{Dataset description}} Additional information from the dataset used in this study is summarized in Extended Data Table \ref{extab1}. The demographic information of the data set is provided in the Extended Data Table \ref{extab3}. Data grouping of different datasets is shown in the Extended data Fig. \ref{figext22}.

\textit{CA dataset -} Arrhythmia data mainly come from CPSC2018 \cite{liu2018open}, CPSC2018-Extra \cite{liu2018open}, PTB \cite{bousseljot1995nutzung}, PBB-XL \cite{liu2018open}, Georgia\footnote[1] {\href{https://physionet.org/content/challenge-2021/1.0.3/}{https://physionet.org/content/challenge-2021/1.0.3/}}, Chapman Shaoxing \cite{zheng202012}, and Ningbo \cite{zheng2020optimal} of the CinC2021 Challenge \cite{goldberger2000physiobank,alday2020classification,reyna2021will}. Georgia (Occidental) was used as external test data to evaluate model performance. In addition, to further evaluate the performance of our model on different population data, SPH \cite{liu2022large} (Asian) was used as external test data to evaluate the performance of AI model. Further, Cardio-valid \cite{sangha2022automated} data was used to evaluate the performance of the model to compare the diagnosis results with cardiologists. First-degree atrioventricular block (AVBI), atrial fibrillation (AF), left bundle branch block (LBBB), right bundle branch block (RBBB), sinus bradycardia (SB), sinus tachycardia (NSR), and sinus rhythm (SA) were used in CA detection tasks. After data screening, CPSC2018, CPSC2018-Extra, PTB, PTB-XL, Chapman-Shaoxing, and Ningbo contained 32,807 ECG records of 10s from 30,688 patients. This data were randomly divided into training sets, internal validation sets, and internal test sets with 9:0.5:0.5. External test data (Georgia, SPH, and Cardio-valid) contained 2871 ECG records from 2871 patients, 16567 from 16460 patients, and 146 from 146 patients, respectively. ECG records in the arrhythmia dataset were standard 12-lead recordings with a sampling frequency of 500Hz to 1000Hz and time of 6 seconds to 1800 seconds. The sampling frequency of most ECG records was 500Hz and the time is 10s. Therefore, we used interp1d\footnote[1]{\href{https://docs.scipy.org/doc/scipy/reference/generated/scipy.interpolate.interp1d.html}{https://docs.scipy.org/doc/scipy/reference/generated/scipy.interpolate.interp1d.html}} to resample the sampling rate of the ECG records to 500Hz. The ECG records in the dataset were clipped with a set length of 5000 (10 seconds $\times$ 500 Hz). ECG records with lengths less than 5000 would be discarded. Then, notch filter\footnote[2]{\href{https://mne.tools/stable/generated/mne.filter.notch_filter.html}{https://mne.tools/stable/generated/mne.filter.notch\_filter.html}} and band pass filter\footnote[3]{\href{https://mne.tools/stable/generated/mne.filter.filter_data.html}{https://mne.tools/stable/generated/mne.filter.filter\_data.html}} were used to filter the clipped ECG records. Finally, the filtered ECG records were drawn into the corresponding ECG image using the ECG plot\footnote[4]{\href{https://pypi.org/project/ecg-plot/}{https://pypi.org/project/ecg-plot/}} in Python (paper speed: 25mm/s, amplitude: 10mm/mV). Each large block of the ECG image was a voltage of 0.5 mV and a duration of 0.2 seconds, respectively.

\textit{AF screening dataset -} Chaoyang is a clinical ECG image dataset obtained during AF screening task with sinus rhythm. The dataset included standard 12-lead ECG image (paper speed: 25mm/s, amplitude: 10mm/mV) in supine state available at Beijing Chaoyang Hospital, Capital Medical University from January 2011 to January 2021. These ECG images were obtained from patient with paroxysmal AF with normal sinus rhythm (YAF) and patient with non-paroxysmal AF with normal sinus rhythm (NAF) who were admitted to these hospitals during the same period. ECG images collected were divided into NAF and YAF. The selection criteria of NAF are as follows. First, paper-based ECG evidence of AF is not found in the records. Second, no previous history of AF. Third, at least one standard 12-lead paper-based ECG of normal sinus rhythm is included in the records. The selection criteria of YAF are as follows. First, paper-based ECG evidence of AF is captured in at least one of the records. Second, a standard 12-lead paper-based ECG of normal sinus rhythm prior to at least one episode of AF is included in the records. In addition, the paper-based ECG of sinus rhythm must be performed before the onset of AF and the interval should be less than 1 month. The exclusion criteria of NAF and YAF are as follows. First, no paper-based ECG of normal sinus rhythm in the records. Second, the records contained a diagnosis of AF and no paper-based ECG evidence of AF in the records. Third, paper-based ECG with severe interference or missing lead. Fourth, the paper-based ECG of cardiac pacing patients. The list of included patients was identified by searching the patient database of Beijing Chaoyang Hospital, Capital Medical University. After that, ECG images were obtained by scanning the paper-based ECG from the records of patient. Finally, a total of 5866 ECG images were collected from 2192 patients with NAF or YAF. The study proposal was reviewed and approved by Ethics Committee of Beijing Chaoyang Hospital, Capital Medical University (No. 2021-ke-555).

Shaoyifu is another clinical dataset obtained from the affiliated Sir Run Run Shaw Hospital of Zhejiang University School of Medicine. Shaoyifu contained 248 ECG images (100 NAF and 148 YAF) from 248 patients. Shaoyifu was used as an external test dataset for the screening task in this study. The study proposal was reviewed and approved by Ethics Committee of Sir Run Run Shaw Hospital, Zhejiang University School of Medicine (No. keyan20211116-30).

{\bfseries{Solutions for analyzing ECG images in the real world}} Lightweight neural network structure, cloud service, image classification algorithm, and so on were introduced to overcome difficulties and challenges such as data dimension difference, interference from the natural environment, and the small number of trainable samples in paper-based ECG images taken by mobile devices. The proposed AI system framework is shown in Extended Data Fig. \ref{figext2}. 

\begin{figure}[ht]
    \centering
    \includegraphics[width=11cm]{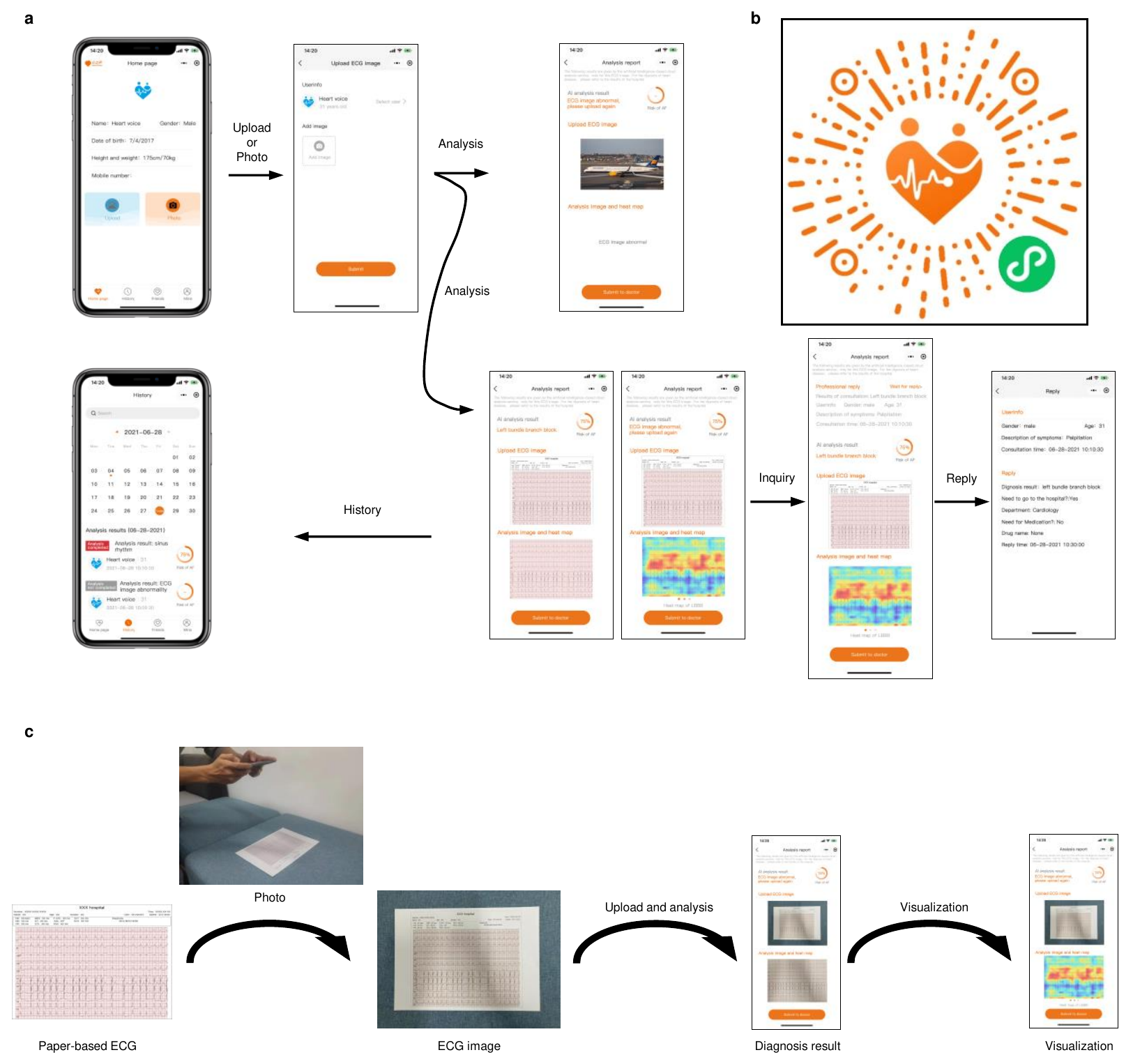}
    \caption{{\bfseries{Overview and workflow details of our system.}} {\bfseries{a}}, an overview of the AI system presented in this study. The interface and analysis flow of our system are shown along with the arrows. {\bfseries{b}}, The QR code of our system can be accessed. Users can scan this QR code via WeChat to access our system. {\bfseries{c}}, The use of our system is visualized. }
    \label{figext1}
\end{figure}

\begin{figure}[htp]
    \centering
    \includegraphics[width=11cm]{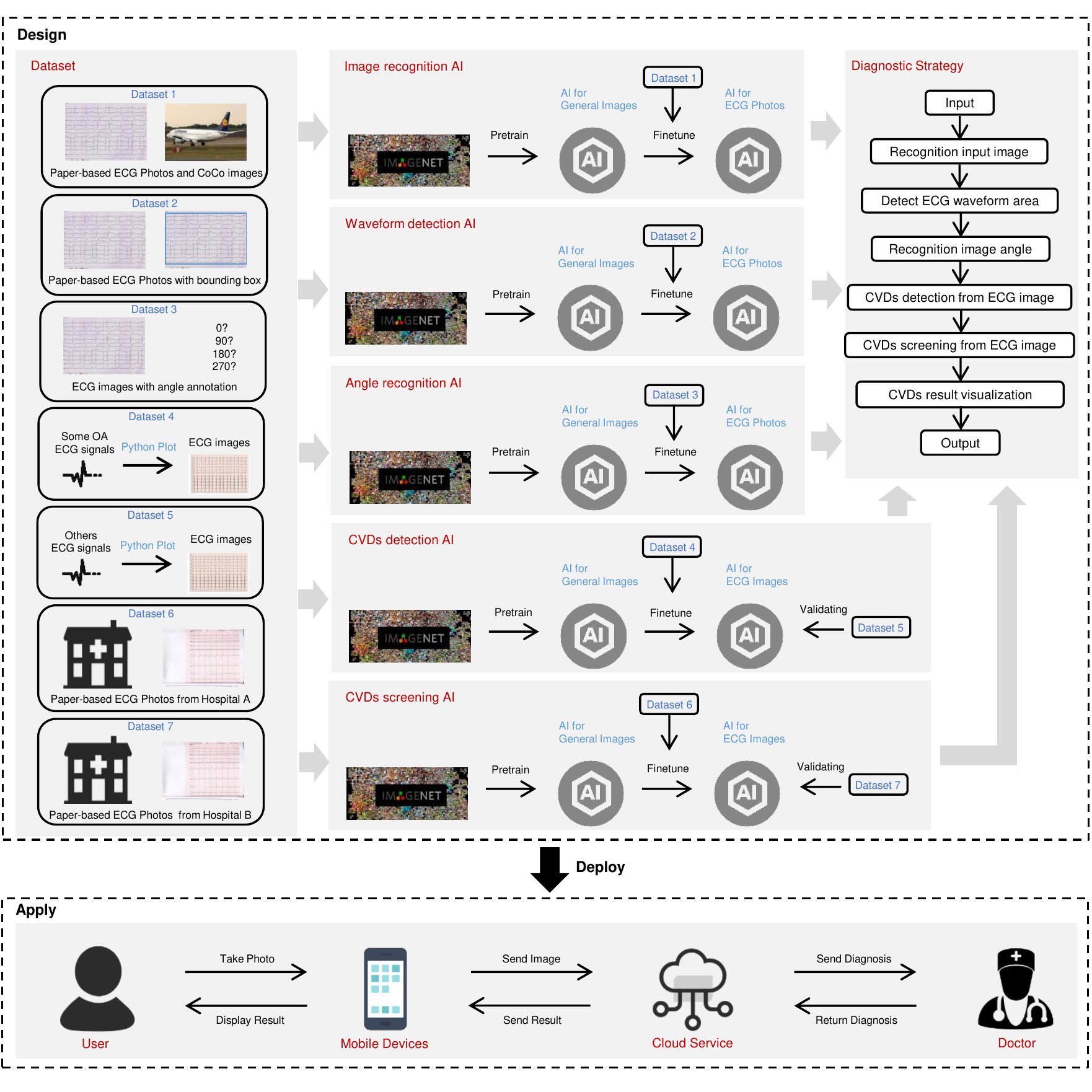}
    \caption{{\bfseries{Overall research and design of the proposed system.}} {\bfseries{Design}}, overview of AI analysis methods and rational formulation of diagnosis strategies. Dataset, the datasets needed to build our model. Image recognition AI, training model for ECG image recognition tasks in order to prevent user error input. Waveform detection AI, training model for waveform detection tasks of ECG image to eliminate the interference of irrelevant information. Angle recognition AI, training model for ECG image angle recognition tasks to correct the angle. CAs detection AI, AI model for detecting CAs from ECG image. CAs screening AI, AI model for screening CAs from ECG image. Diagnostic Strategy, The reasonable decision effectively solves the challenges and difficulties of ECG analysis in the real world. More critical areas are provided to the detection and screening model. {\bfseries{Apply}}, showing the details and workflow of our system.}
    \label{figext2}
\end{figure}

\begin{figure}[htp]
    \centering
    \includegraphics[width=11cm]{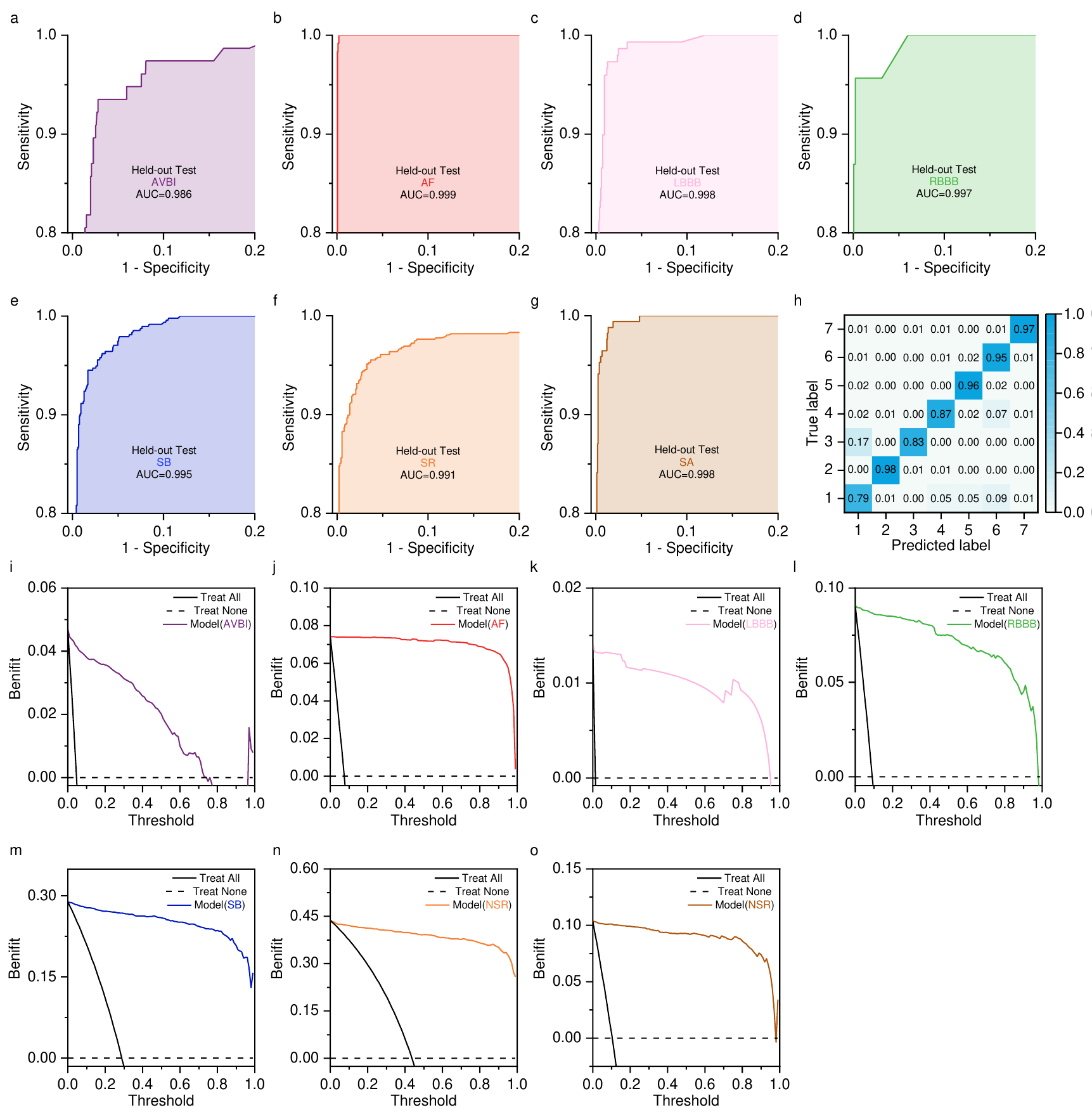}
    \caption{{\bfseries{Performance of the CA detection model on hold-out dataset.}} {\bfseries{a-g}}, ROC curves of different diseases for detecting CAs from ECG images in hold-out test dataset (n = 1640 ECG images from 1535 patients). {\bfseries{h}}, confusion matrix of CAs detection model on hold-out test dataset. {\bfseries{i-o}}, DCA of different diseases for detecting CAs from ECG images in hold-out test dataset.}
    \label{figext3}
\end{figure}

\begin{figure}[htp]
    \centering
    \includegraphics[width=11cm]{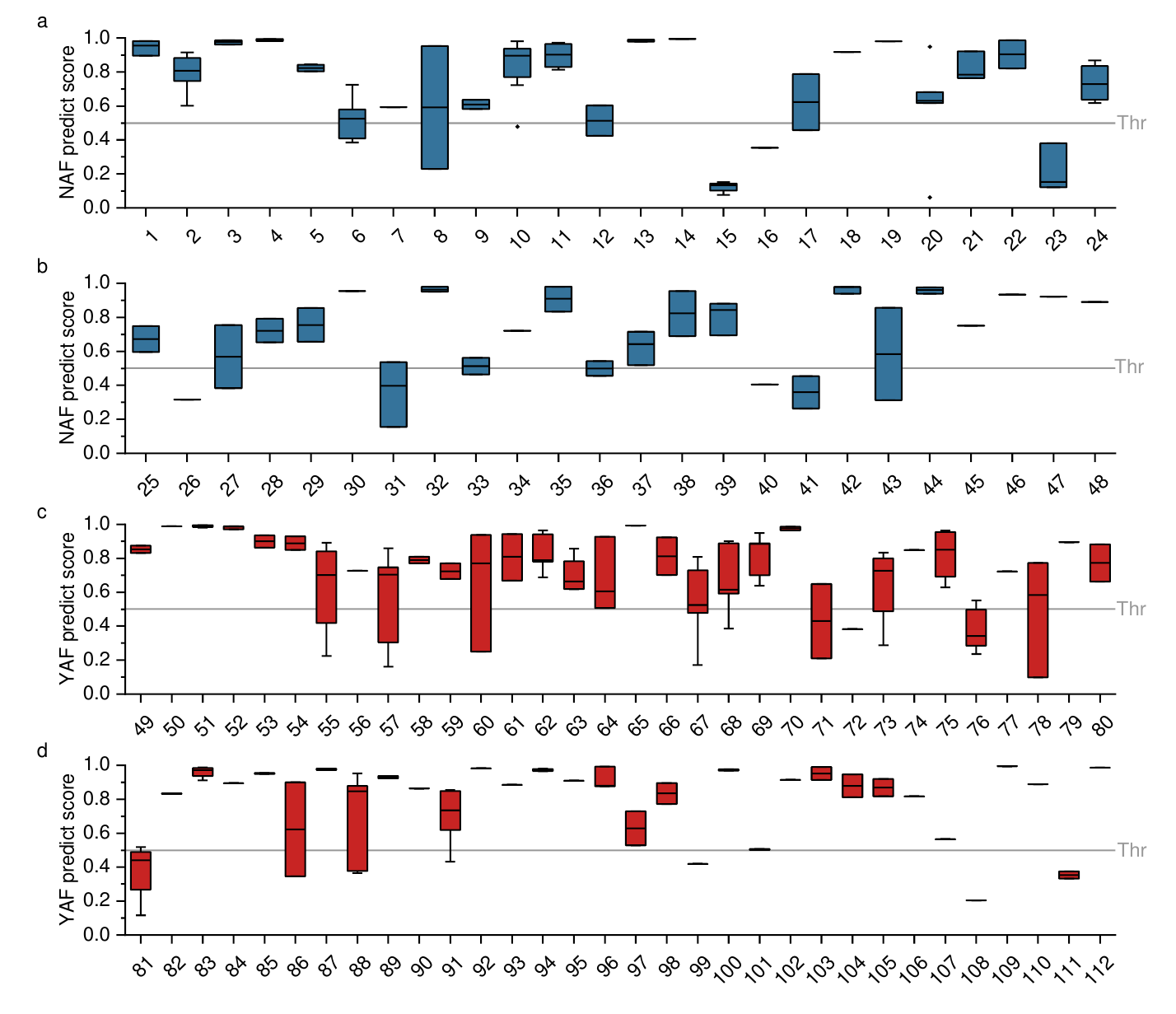}
    \caption{{\bfseries{Distribution of AF risk scores in patients with multiple ECG images (n = 290 ECG images from 112 patients, 48(NAF, a, b) vs 64(YAF, c, d)).}} These graphs show boxplots of risk scores for AF. On each box, the center line represents the median, and the bottom and top edges of the box represent the 25th and 75th percentiles, respectively. The extension beyond the box represents 1.5 times the interquartile range. Outliers are represented using diamonds. Thresholds are shown by gray horizontal lines.}
    \label{figext4}
\end{figure}

\begin{figure}[htp]
    \centering
    \includegraphics[width=11cm]{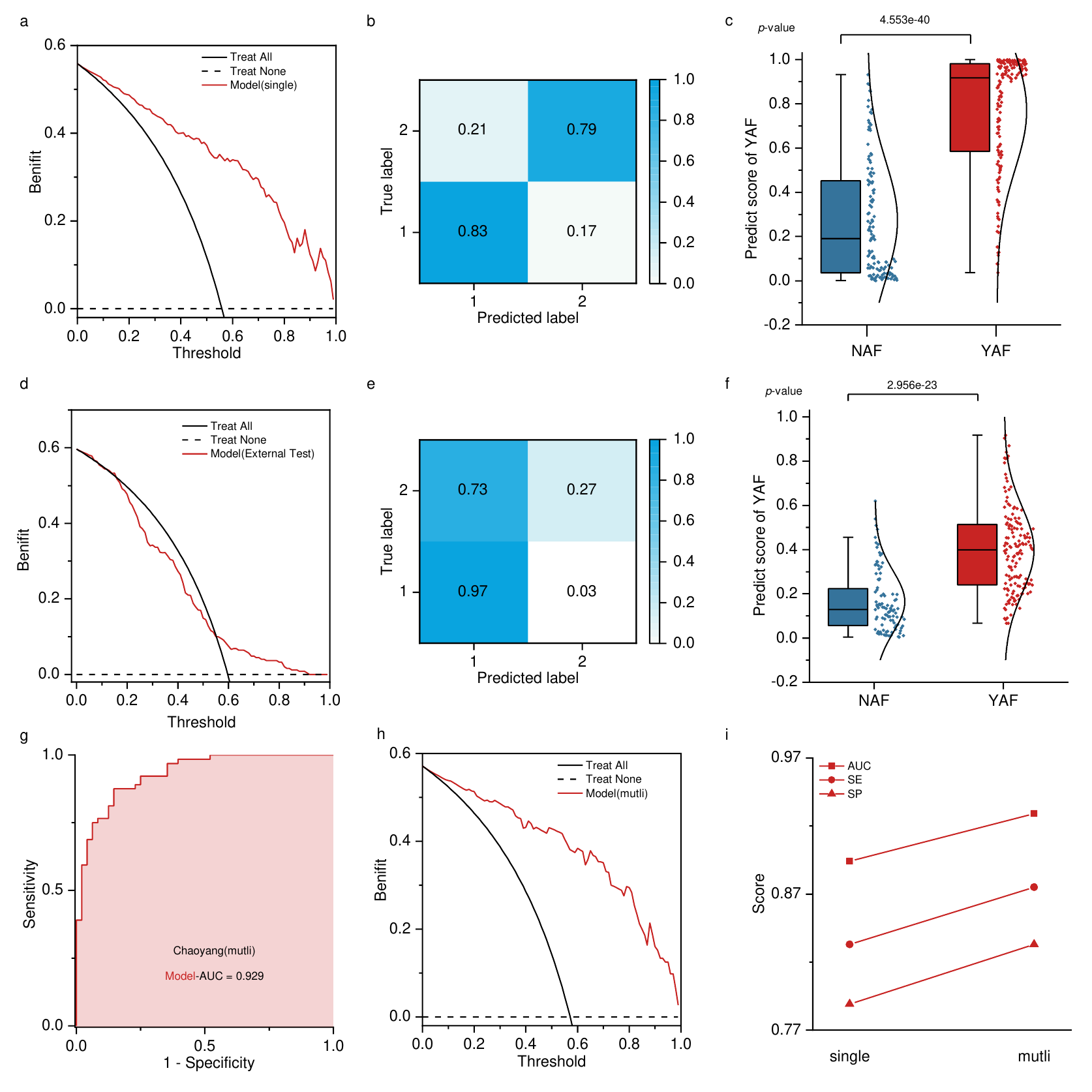}
    \caption{{\bfseries{Performance of the AF screening model in the Chaoyang dataset.}} {\bfseries{a}}, Decision curve for screening CAs from ECG images in Chaoyang dataset. {\bfseries{b}}, confusion matrix of CAs screening model on Chaoyang dataset (the threshold is 0.5). {\bfseries{c}}, The boxplots and normal distribution curves on the Chaoyang dataset.{\bfseries{d}}, Decision curve for screening CAs from ECG images in Shaoyifu dataset. {\bfseries{e}}, confusion matrix of CAs screening model on Shaoyifu dataset (the threshold is 0.5). {\bfseries{f}}, The boxplots and normal distribution curves on the Shaoyifu dataset. {\bfseries{g}}, ROC curve after ensemble multiple ECG image predictions for each patient. {\bfseries{h}}, Decision curve after ensemble multiple ECG image predictions for each patient. {\bfseries{i}}, A comparison result of the prediction based on the ECG image (single) and patient (mutli) in the case of multiple ECG images provided by one patient.}
    \label{figext5}
\end{figure}

\begin{figure}[htp]
    \centering
    \includegraphics[width=11cm]{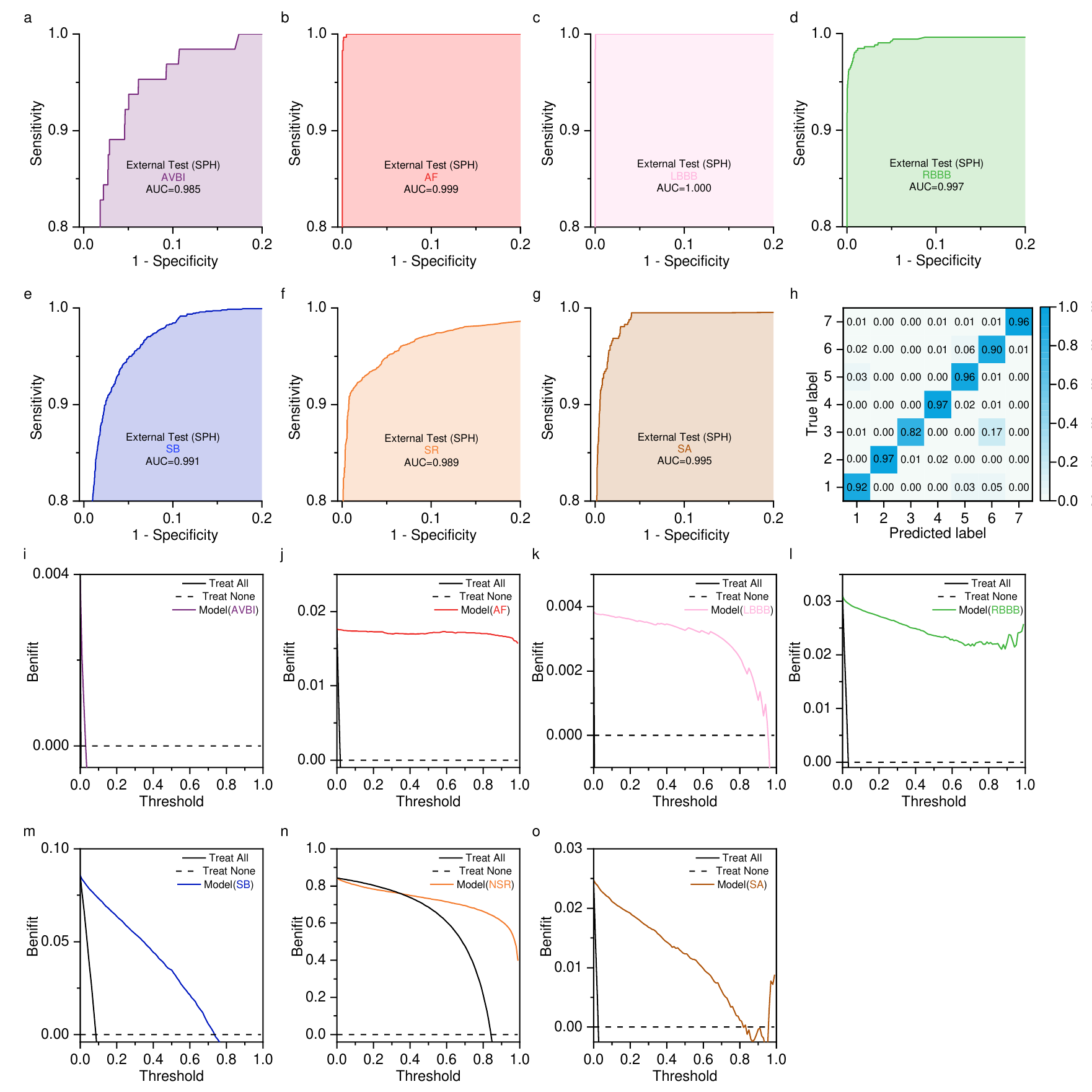}
    \caption{{\bfseries{Performance of the CA detection model on SPH dataset.}} {\bfseries{a-g}}, ROC curve of different diseases for detecting CAs from ECG images in SPH dataset (n = 16567 ECG images from 16460 patients). {\bfseries{h}}, confusion matrix of CAs detection model on SPH dataset. {\bfseries{i-o}}, DCA of different diseases for detecting CAs from ECG images in SPH dataset.}
    \label{figext6}
\end{figure}

\begin{figure}[htp]
    \centering
    \includegraphics[width=11cm]{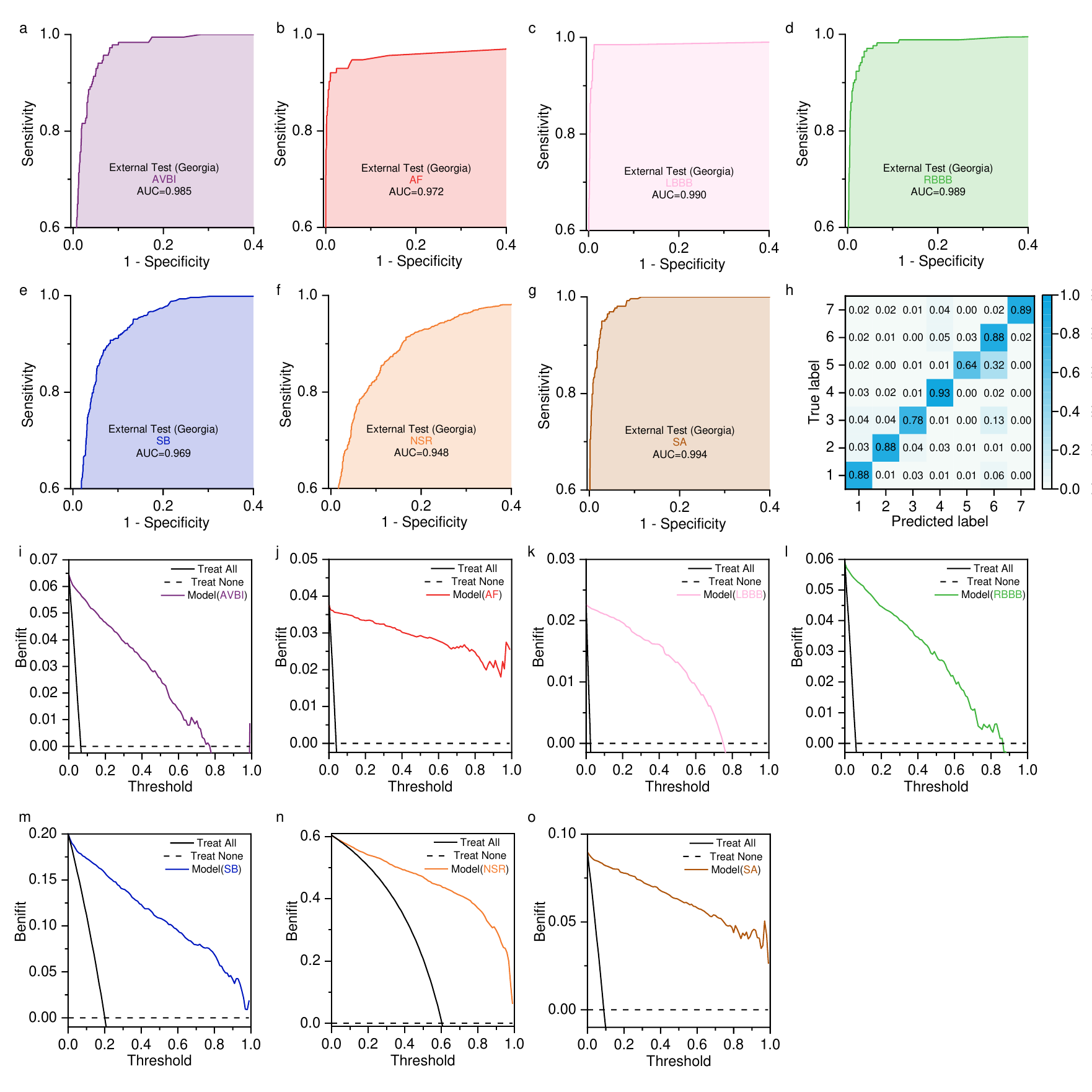}
    \caption{{\bfseries{Performance of the CA detection model on Georgia dataset.}} {\bfseries{a-g}}, ROC curve of different diseases for detecting CAs from ECG images in Georgia dataset (n = 2871 ECG images from 2871 patients). {\bfseries{h}}, confusion matrix of CAs detection model on Georgia dataset. {\bfseries{i-o}}, DCA of different diseases for detecting CAs from ECG images in Georgia dataset.}
    \label{figext7}
\end{figure}

\begin{figure}[htp]
    \centering
    \includegraphics[width=11cm]{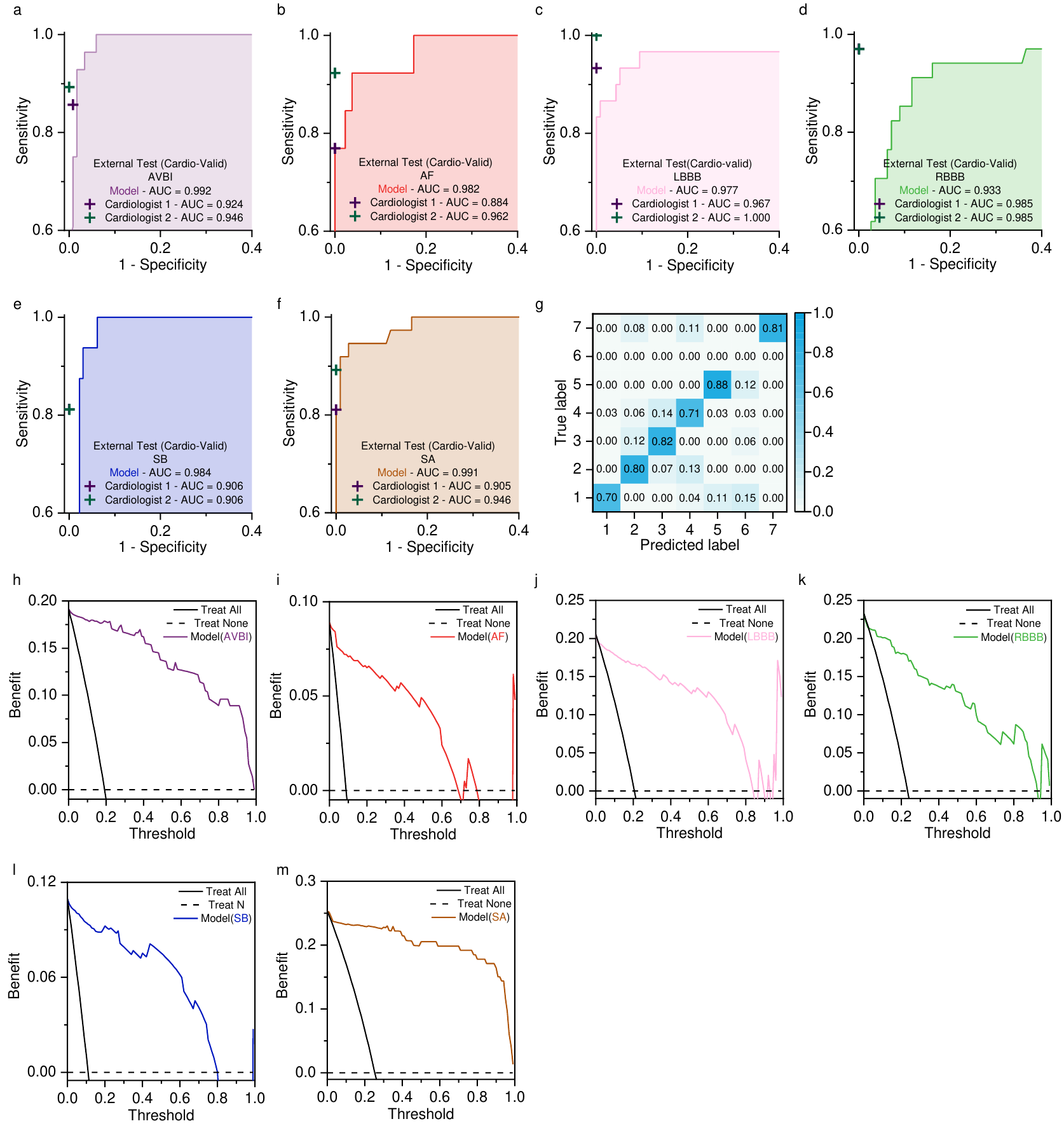}
    \caption{{\bfseries{Performance of the CA detection model on Cardio-valid dataset.}} {\bfseries{a-f}}, ROC curve of different diseases for detecting CAs from ECG images in Cardio-valid dataset (n = 146 ECG images from 146 patients). {\bfseries{g}}, confusion matrix of CAs detection model on Cardio-valid dataset. {\bfseries{h-m}}, DCA of different diseases for detecting CAs from ECG images in Cardio-valid dataset.}
    \label{figext8}
\end{figure}

\begin{figure}[htp]
    \centering
    \includegraphics[width=11cm]{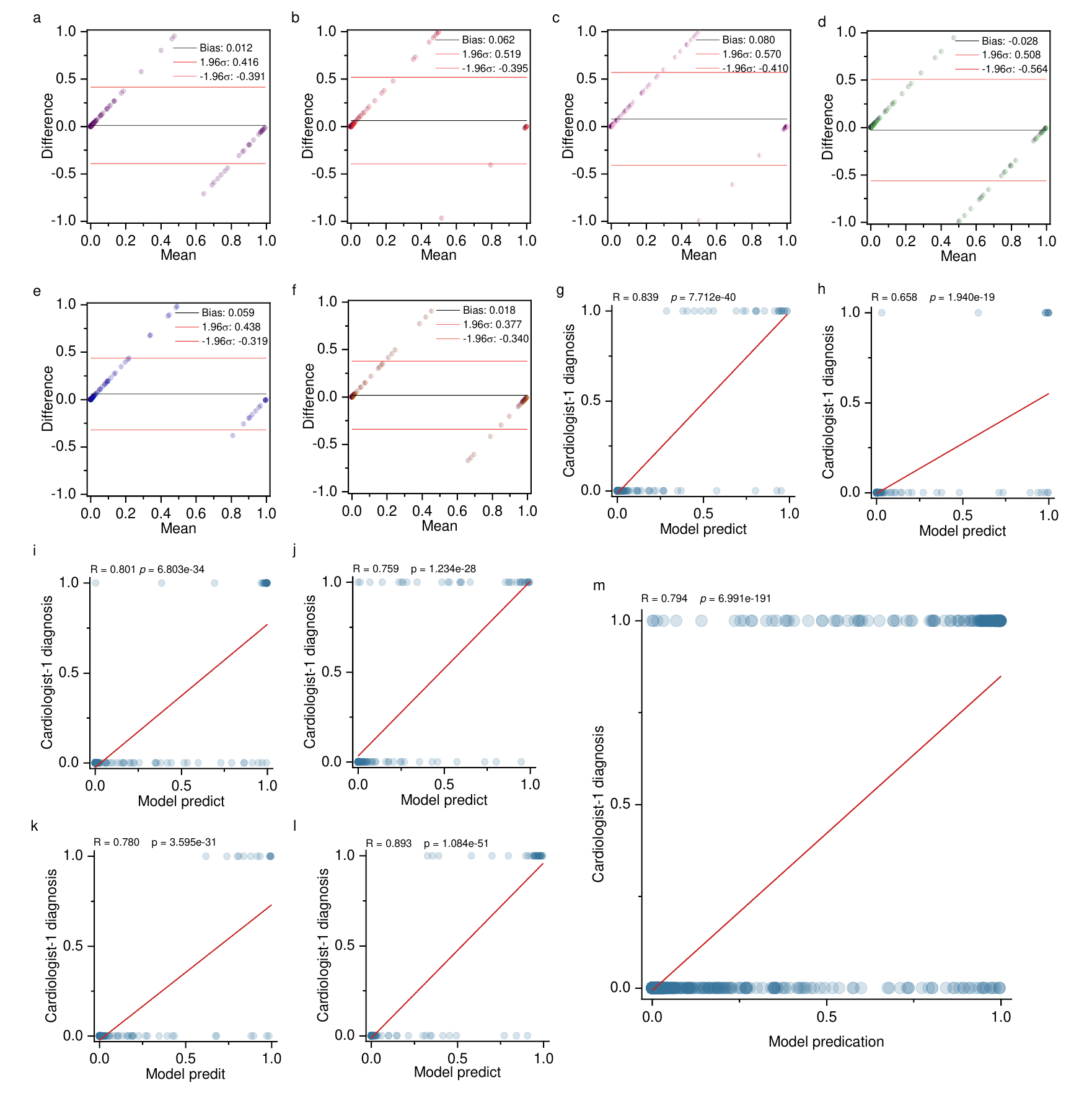}
    \caption{{\bfseries{Compared diagnostic results of CAs detection models and cardiologists 1.}} {\bfseries{a-f}}, Bland-Altman plot for detection CAs using diagnosis from our model and cardiologists 1 on different arrhythmias. {\bfseries{g-l}},correlation plot for detection CAs using diagnosis from our model and cardiologists 1 on different arrhythmias. {\bfseries{m}}, correlation plot for detection CAs using diagnosis from our model and cardiologists 1 on Cardio-valid dataset.}
    \label{figext9}
\end{figure}

\begin{figure}[htp]
    \centering
    \includegraphics[width=11cm]{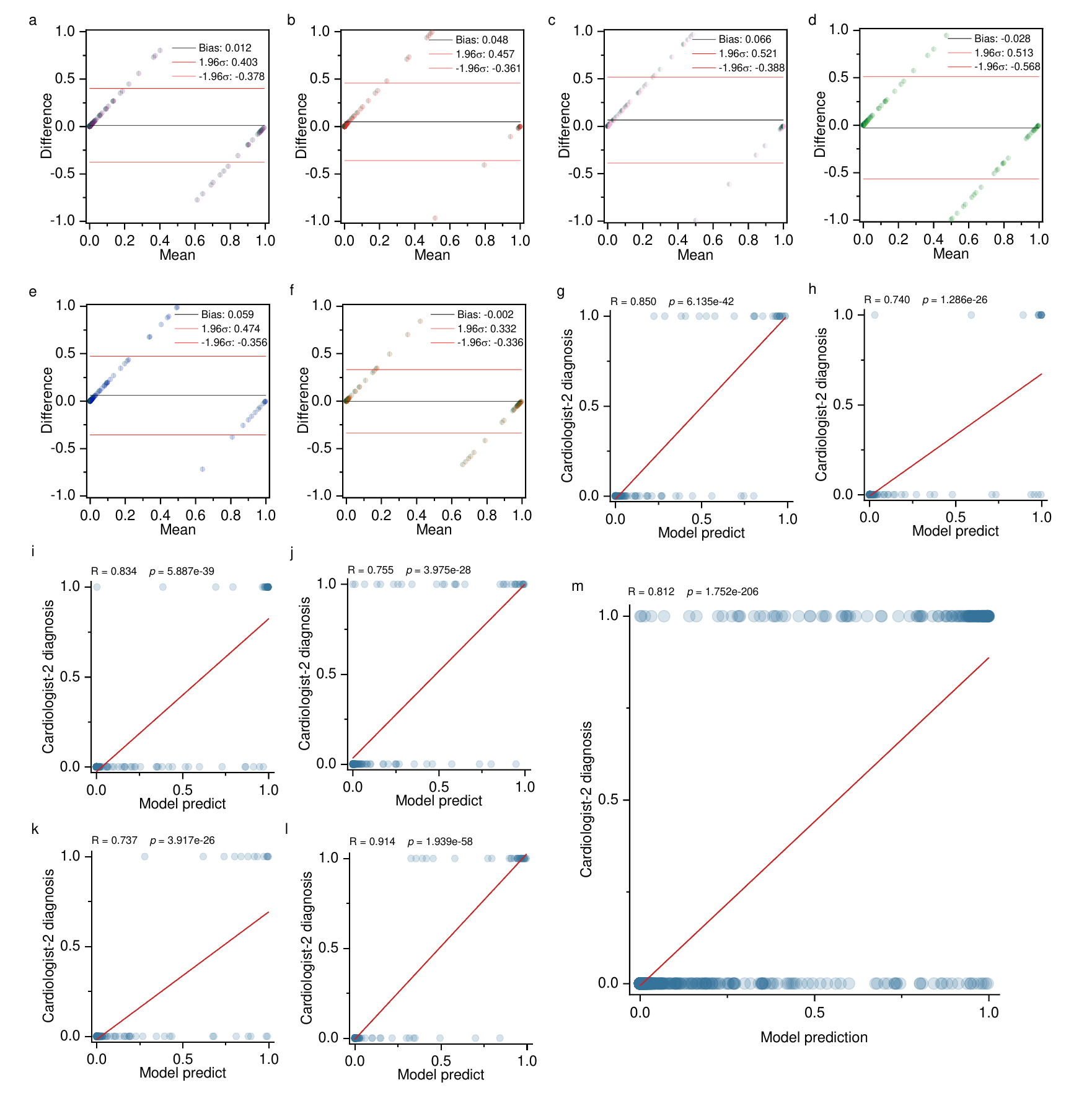}
    \caption{{\bfseries{Compared diagnostic results of CAs detection models and cardiologists 2.}} {\bfseries{a-f}}, Bland-Altman plot for detection CAs using diagnosis from our model and cardiologists 2 on different arrhythmias. {\bfseries{g-l}}, correlation plot for detection CAs using diagnosis from our model and cardiologists 1 on different arrhythmias. {\bfseries{m}}, correlation plot for detection CAs using diagnosis from our model and cardiologists 1 on Cardio-valid dataset.}
    \label{figext10}
\end{figure}

\begin{figure}[htp]
    \centering
    \includegraphics[width=11cm]{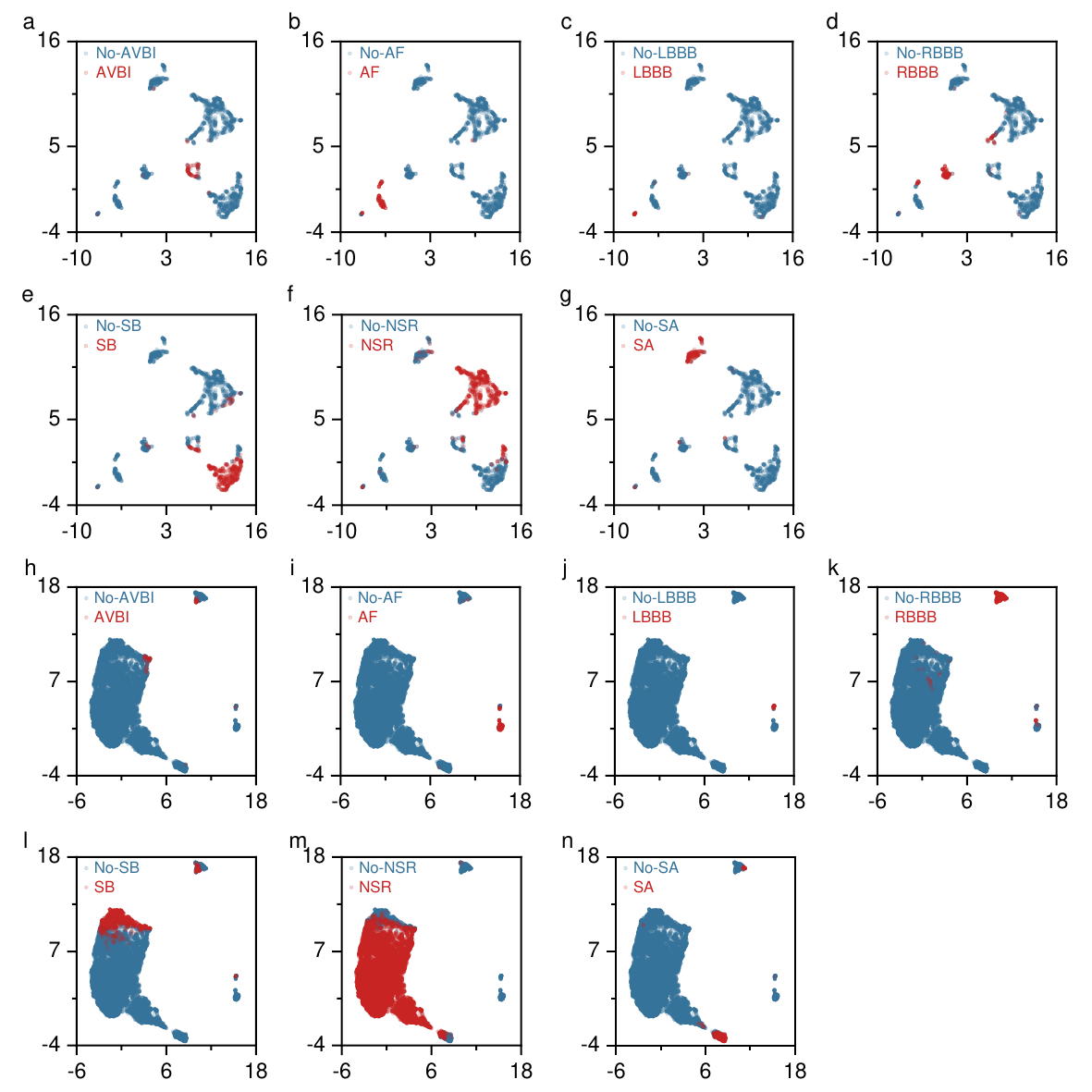}
    \caption{{\bfseries{Scatter plot distribution after UMAP dimension reduction.}} {\bfseries{a-g}}, feature distributed of AVBI, AF, LBBB, RBBB, SB, NSR, and SA from our model in the held-out dataset. {\bfseries{h-n}}, feature distributed of AVBI, AF, LBBB, RBBB, SB, NSR, and SA from our model in the SPH dataset.}
    \label{figext11}
\end{figure}

\begin{figure}[htp]
    \centering
    \includegraphics[width=11cm]{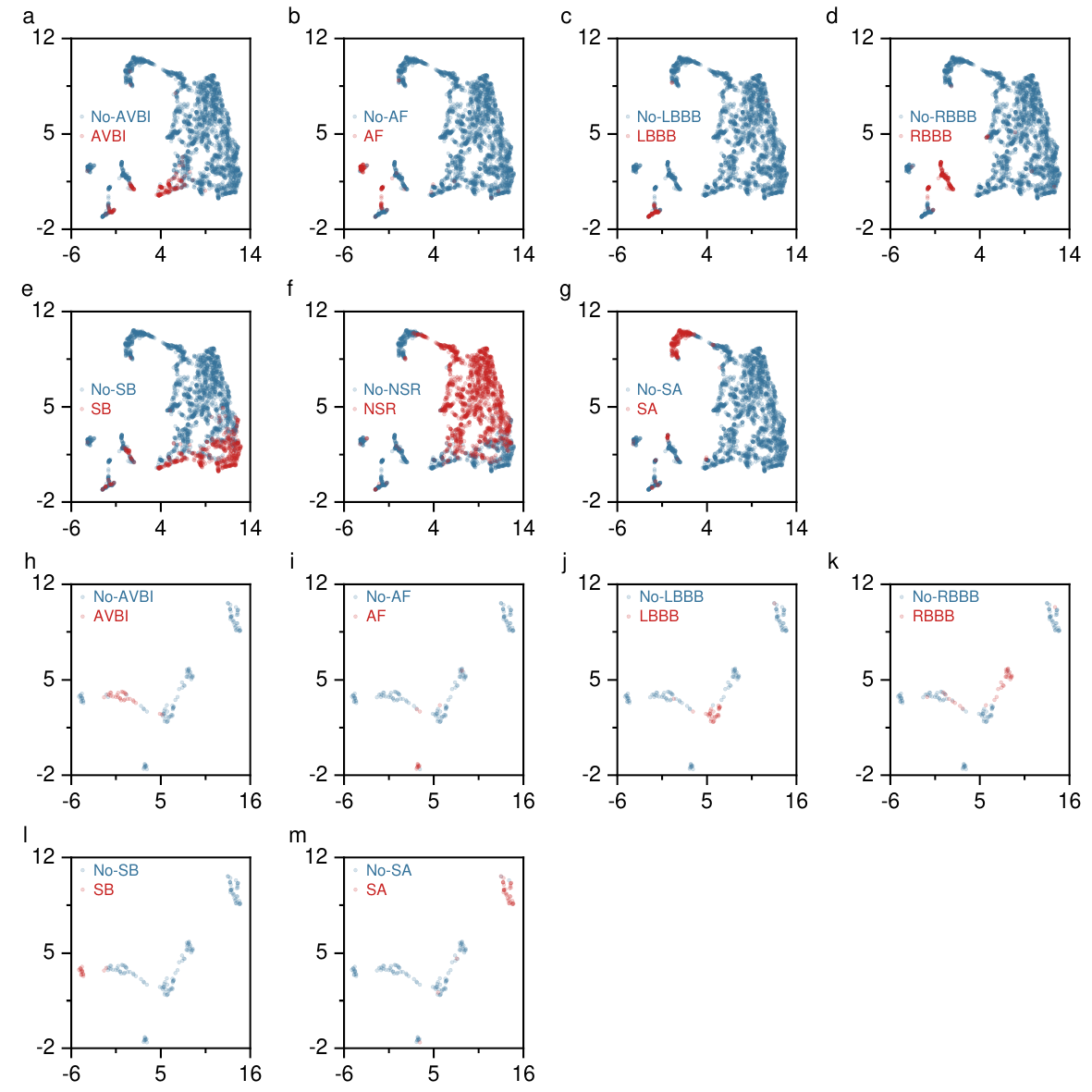}
    \caption{{\bfseries{Scatter plot distribution after UMAP dimension reduction.}} {\bfseries{a-g}}, feature distributed of AVBI, AF, LBBB, RBBB, SB, NSR, and SA from our model in the Georgia dataset. {\bfseries{h-m}}, feature distributed of AVBI, AF, LBBB, RBBB, SB, NSR, and SA from our model in the Cardio-valid dataset.}
    \label{figext12}
\end{figure}

\begin{figure}[htp]
    \centering
    \includegraphics[width=11cm]{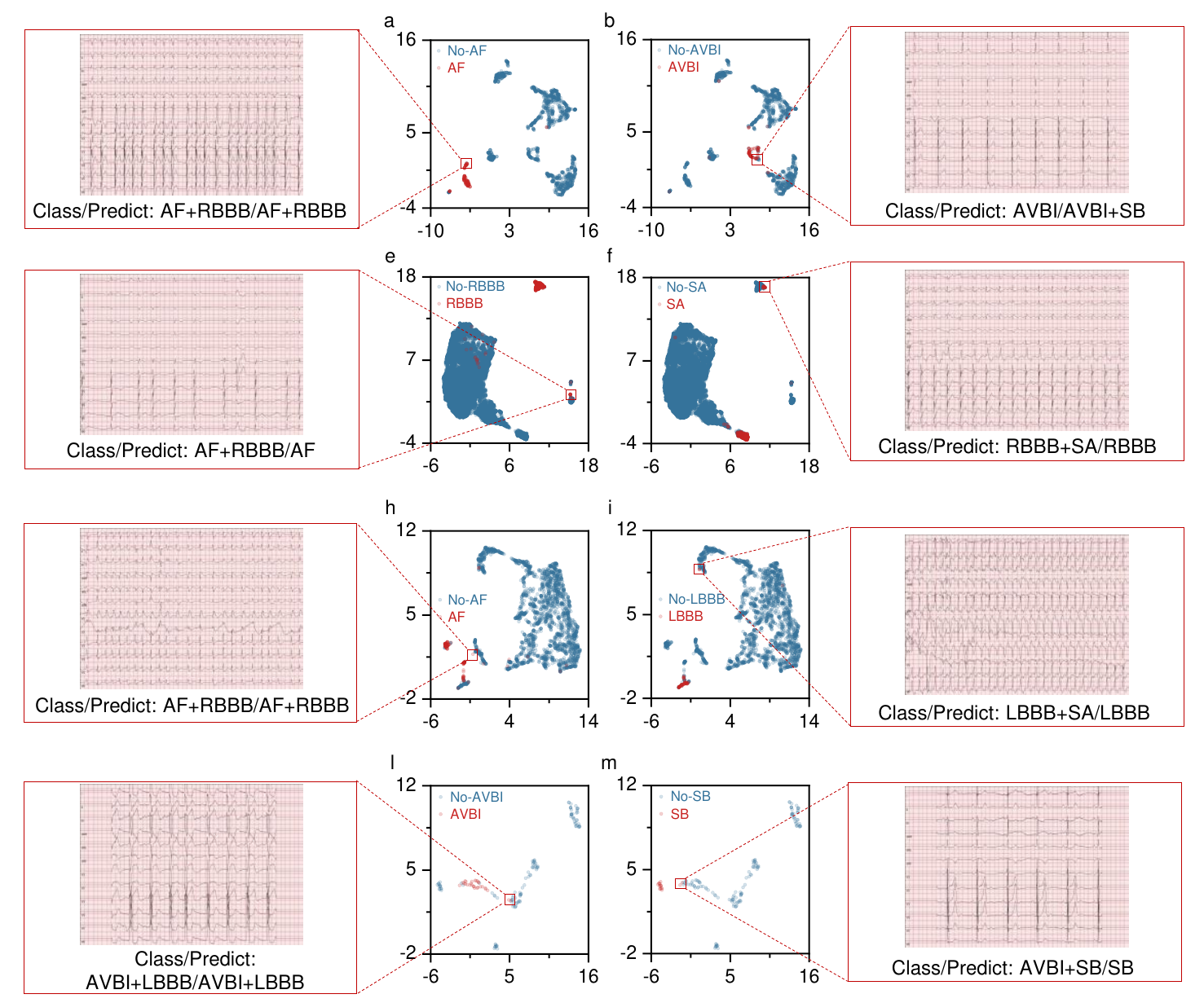}
    \caption{{\bfseries{Visualization of outliers in feature distribution scatter plots.}} The small red box shows where the outliers are. The large red box shows the ECG image represented by outliers. The ground truth and predicted results are shown below the image.}
    \label{figext13}
\end{figure}

\begin{figure}[htp]
    \centering
    \includegraphics[width=11cm]{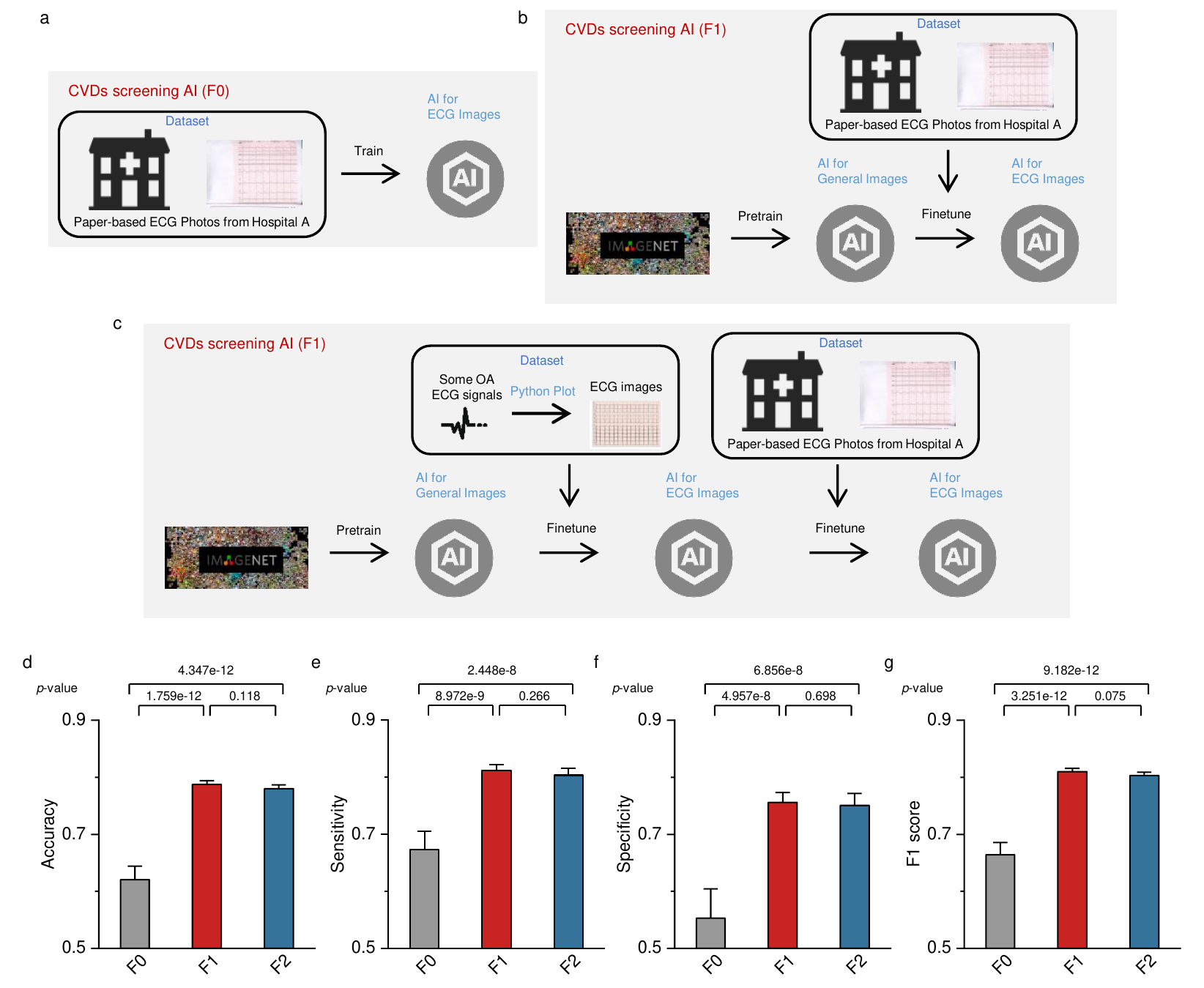}
    \caption{{\bfseries{Results of two-stage fine-tuning ablation experiments.}} {\bfseries{a}}, the training way of CAs screening model without transfer learning. {\bfseries{b}}, the training way of CAs screening model with one-stage fine-tuning transfer learning. {\bfseries{c}}, the training way of CAs screening model with two-stage fine-tuning transfer learning. {\bfseries{d-g}}, Comparison between F0, F1, and F2 when applied to the tasks of AF screening task. The AI model after transfer learning is better than without transfer learning. Compared with one-stage fine-tuning, the performance of two-stage fine-tuning is not significantly improved. Moreover, there was no statistical difference between single-stage and two-stage fine-tuning. The error bars represent the 95\% CI.}
    \label{figext14}
\end{figure}

\begin{figure}[htp]
    \centering
    \includegraphics[width=11cm]{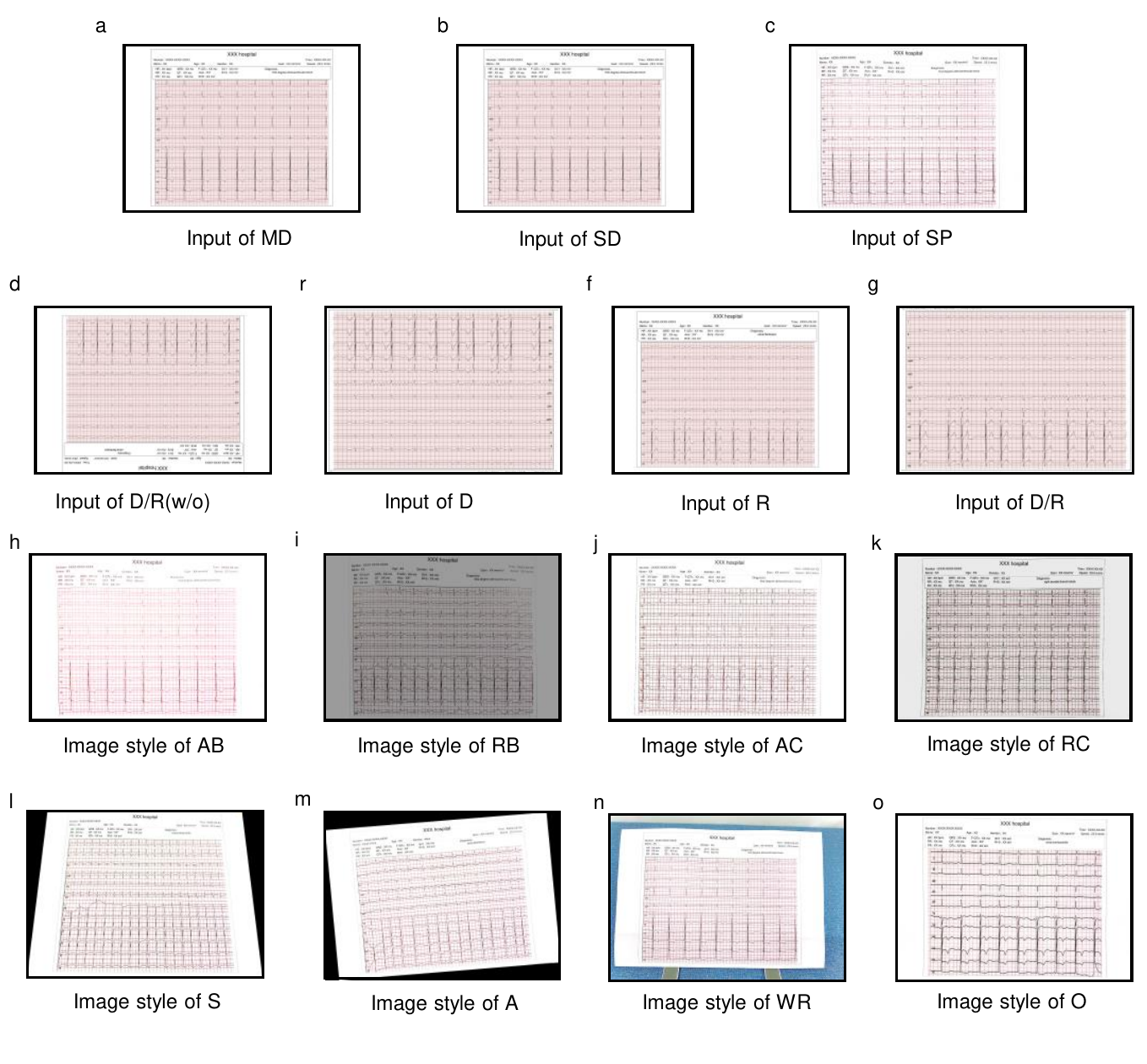}
    \caption{{\bfseries{The style of the ECG image used in this study.}} {\bfseries{a-g}}, examples of different inputs. MD, the prediction results of our model are based on digital data. SD, the prediction results of our system are based on digital data. SP, the prediction results of our system are based on photo data. D/R(w/o), CAs detection and screening were performed directly on the input images without processing. D, CAs detection and screening were performed on the images after waveform region detection. R, CAs detection and screening were performed on the images after angle-corrected. D, CAs detection and screening were performed on the images after waveform region detection and angle-corrected. {\bfseries{h-o}}, ECG images style were obtained under different imaging conditions. AB, add brightness. RB, reduce brightness. AC, add contrast. RC, reduce contrast. S, slant. A, add angle. WR, window ratio. O, origin. }
    \label{figext15}
\end{figure}

\begin{figure}[htp]
    \centering
    \includegraphics[width=11cm]{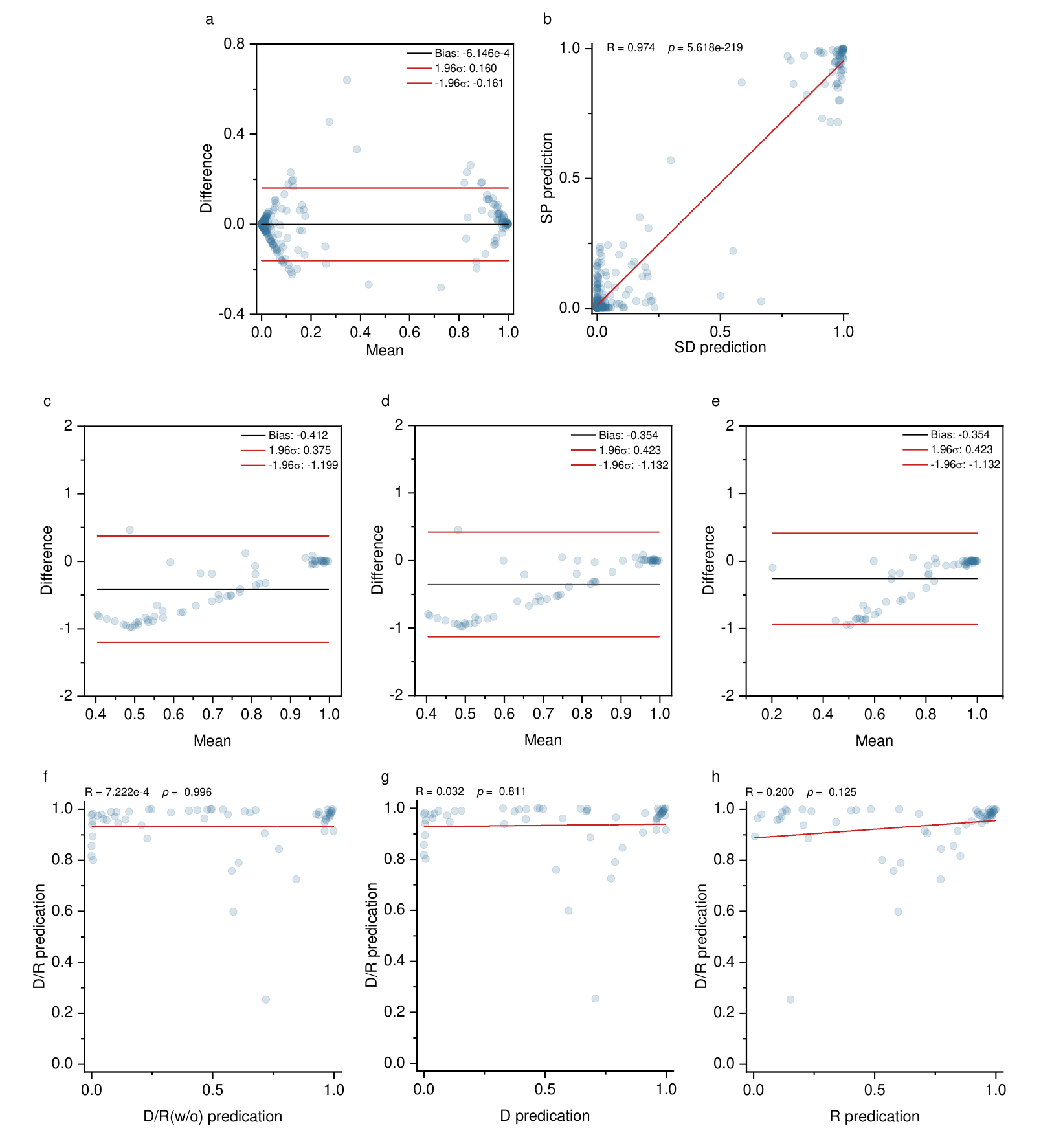}
    \caption{{\bfseries{Consistency and relevance between model and system.}} {\bfseries{a}}, Bland-Altman plot based on photo data (SD and SP). {\bfseries{b}}, correlation between high-quality data (SD) and low-quality data (SP) in the system. {\bfseries{c-e}}, Bland-Altman plots of D/R with D/R(w/o), D, and R, respectively.. {\bfseries{f-h}}, correlation of D/R with D/R(w/o), D, and R, respectively.}
    \label{figext15-16}
\end{figure}

\begin{figure}[htp]
    \centering
    \includegraphics[width=11cm]{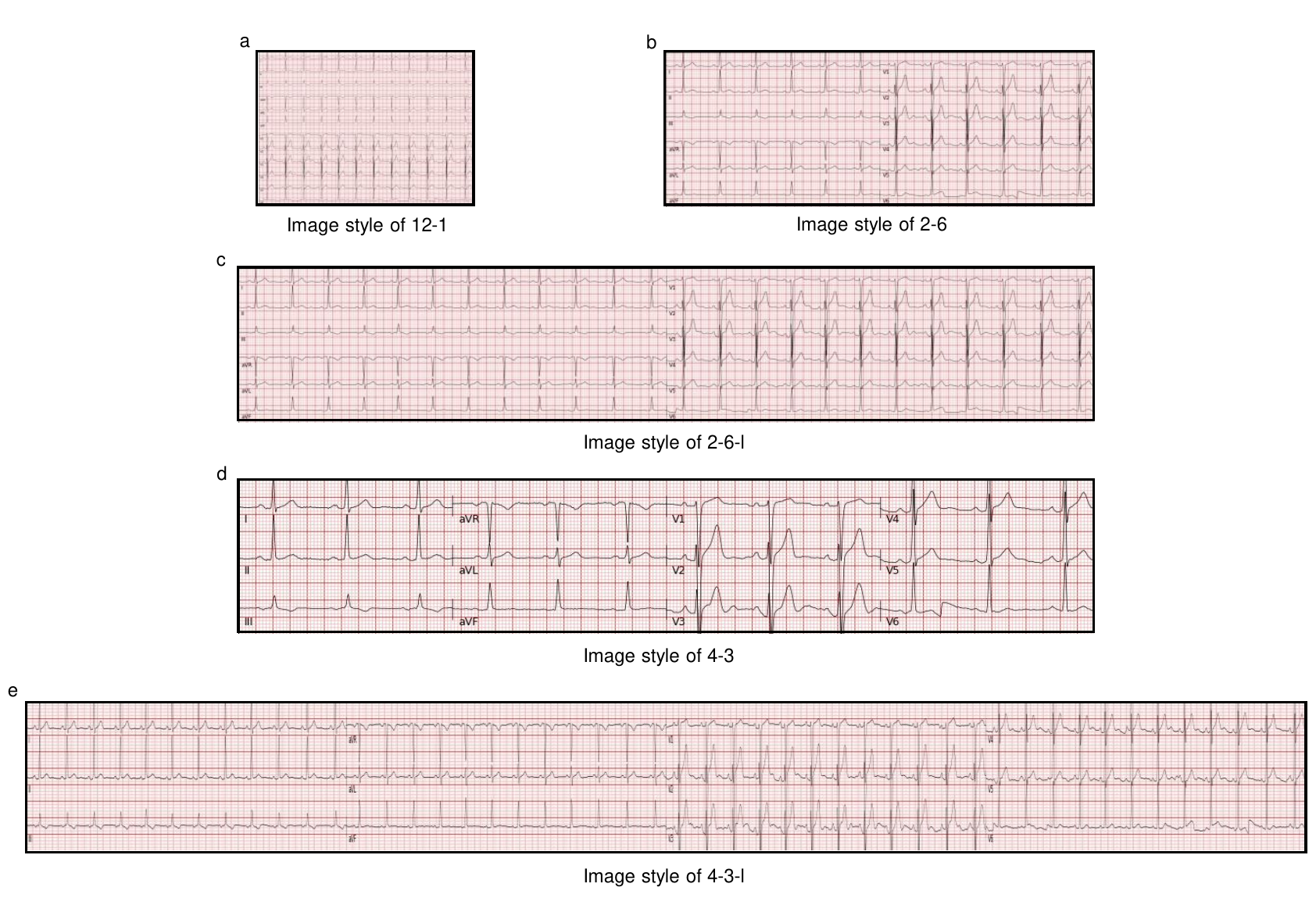}
    \caption{{\bfseries{The style of the ECG image used in this study.}} {\bfseries{a-e}}, 2-6-l, 12-lead ECG signals of 10s are arranged in two columns and six rows. 2-6, 12-lead ECG signals of 5s are arranged in 2 columns and 6 rows. 12-1, 12-lead ECG signals of 10s are arranged in 12 columns and 1 rows. 4-3, 12-lead ECG signals of 2.5s are arranged in 4 columns and 3 rows. 4-3-l, 12-lead ECG signals of 10s are arranged in 4 columns and 3 rows.}
    \label{figext16}
\end{figure}

\begin{figure}[htp]
    \centering
    \includegraphics[width=11cm]{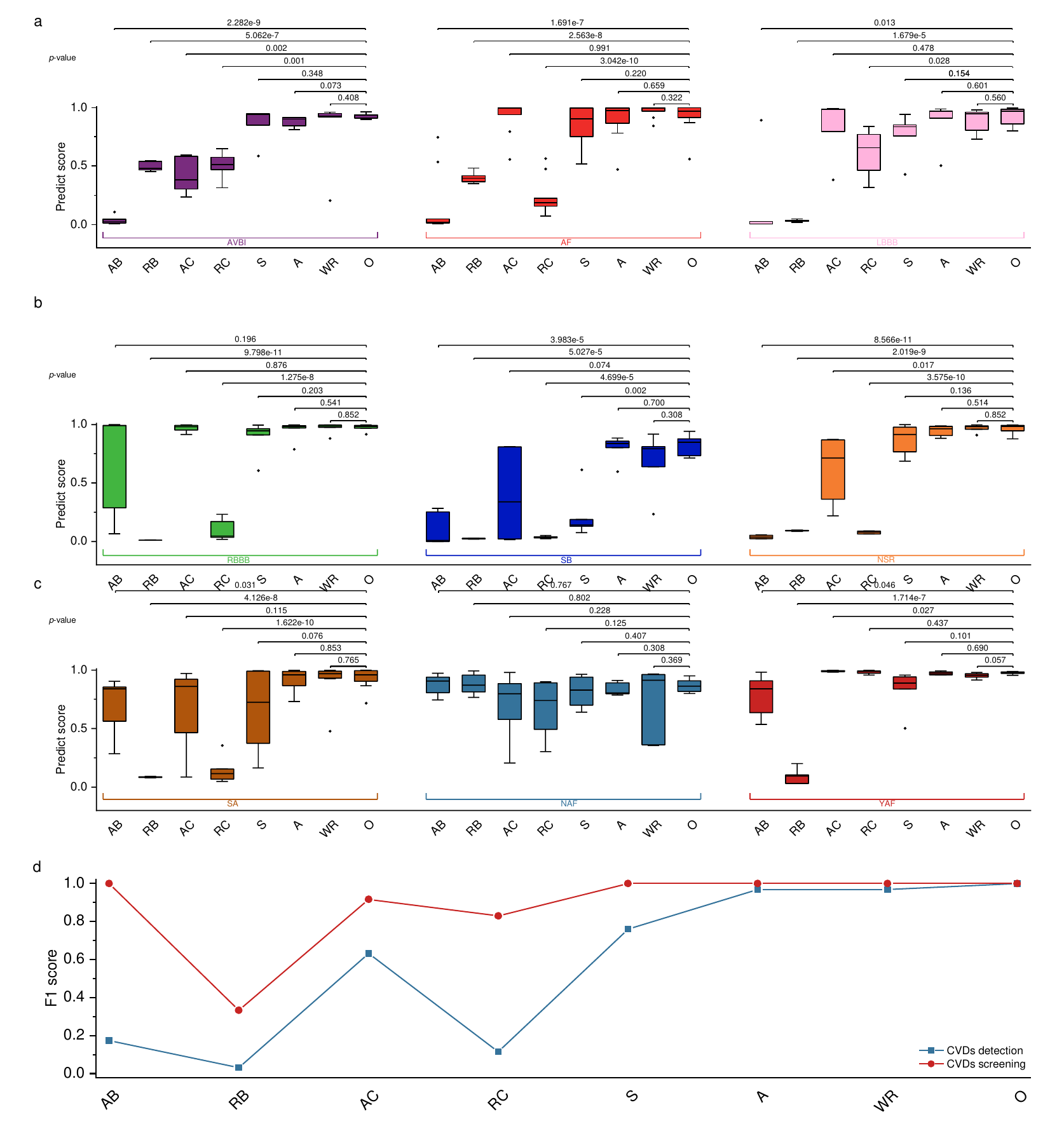}
    \caption{{\bfseries{The influence of imaging condition variation on predictive performance.}} {\bfseries{a-c}}, boxplots of predict results for different imaging condition. The different diseases represented by the different colors are shown under the boxplots. On each box, the center line represents the median, and the bottom and top edges of the box represent the 25th and 75th percentiles, respectively. The extension beyond the box represents 1.5 times the interquartile range. Use diamonds to represent outliers. The p values are shown above the boxplots. {\bfseries{d}}, F1 score curve under different imaging conditions. AVBI, first degree atrioventricular block. AF, atrial fibrillation. LBBB, left bundle branch block. RBBB, right bundle branch block. SB, sinus bradycardia. NSR, sinus rhythm. SA, sinus tachycardia. NAF, sinus rhythm without AF. YAF, sinus rhythm before AF. The \textit{p} values are shown above the boxplot.}
    \label{figext17}
\end{figure}

\begin{figure}[htp]
    \centering
    \includegraphics[width=11cm]{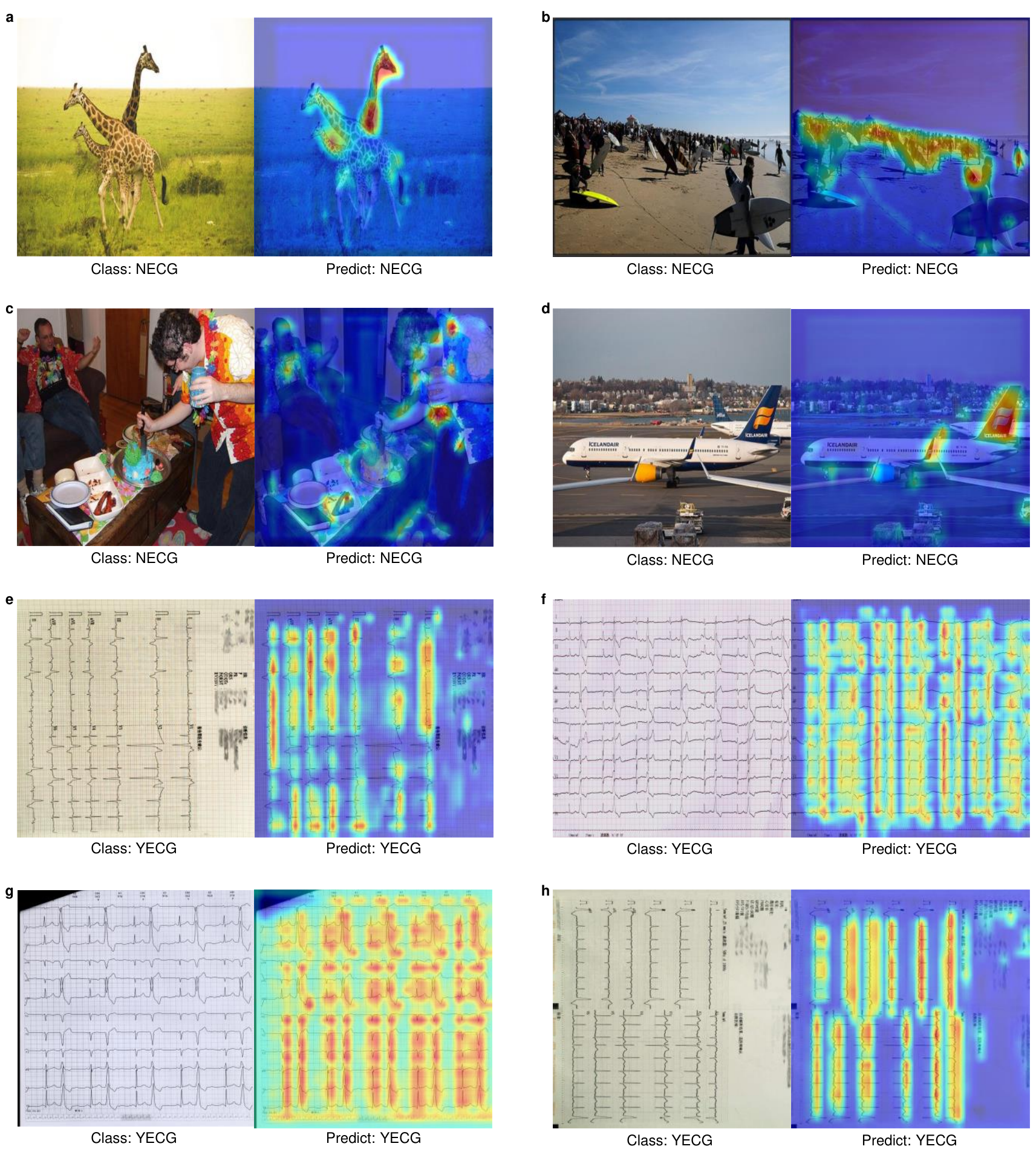}
    \caption{{\bfseries{Results visualization of ECG image recognition task.}} {\bfseries{a-d}}, recognition results of ECG images. {\bfseries{e-h}}, recognition results of ECG images. The original image is shown on the left. The results of the contribution score of the pixels in the ECG image are shown on the right. Real and predicted categories are labeled below the image.}
    \label{figext18}
\end{figure}

\begin{figure}[htp]
    \centering
    \includegraphics[width=11cm]{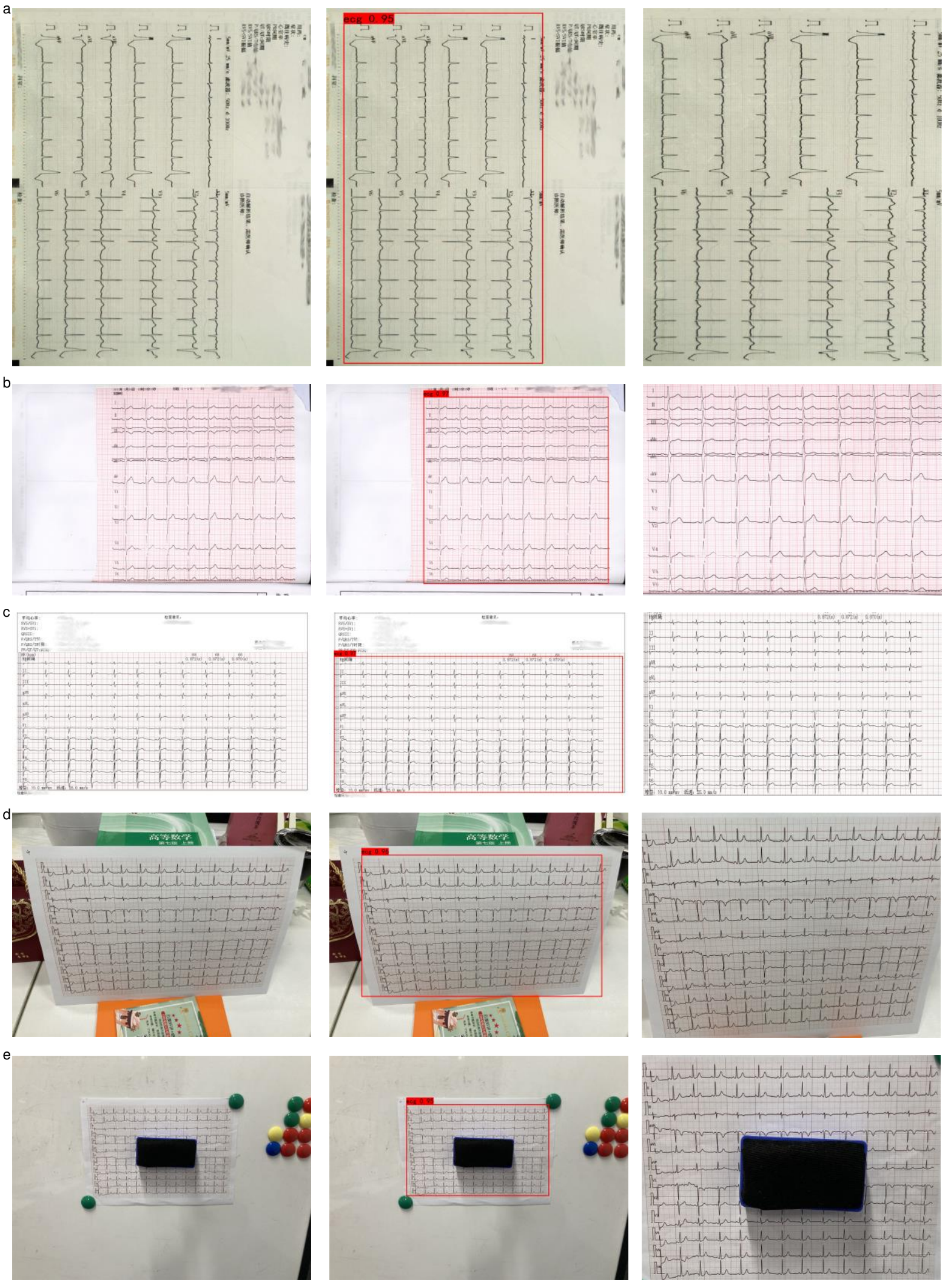}
    \caption{{\bfseries{Results visualization of waveform detection task.}} The original image is shown on the far left. The detection results of our model are shown in the middle. The waveform region obtained after clipping the detection frame area is displayed at the far right.}
    \label{figext19}
\end{figure}

\begin{figure}[htp]
    \centering
    \includegraphics[width=11cm]{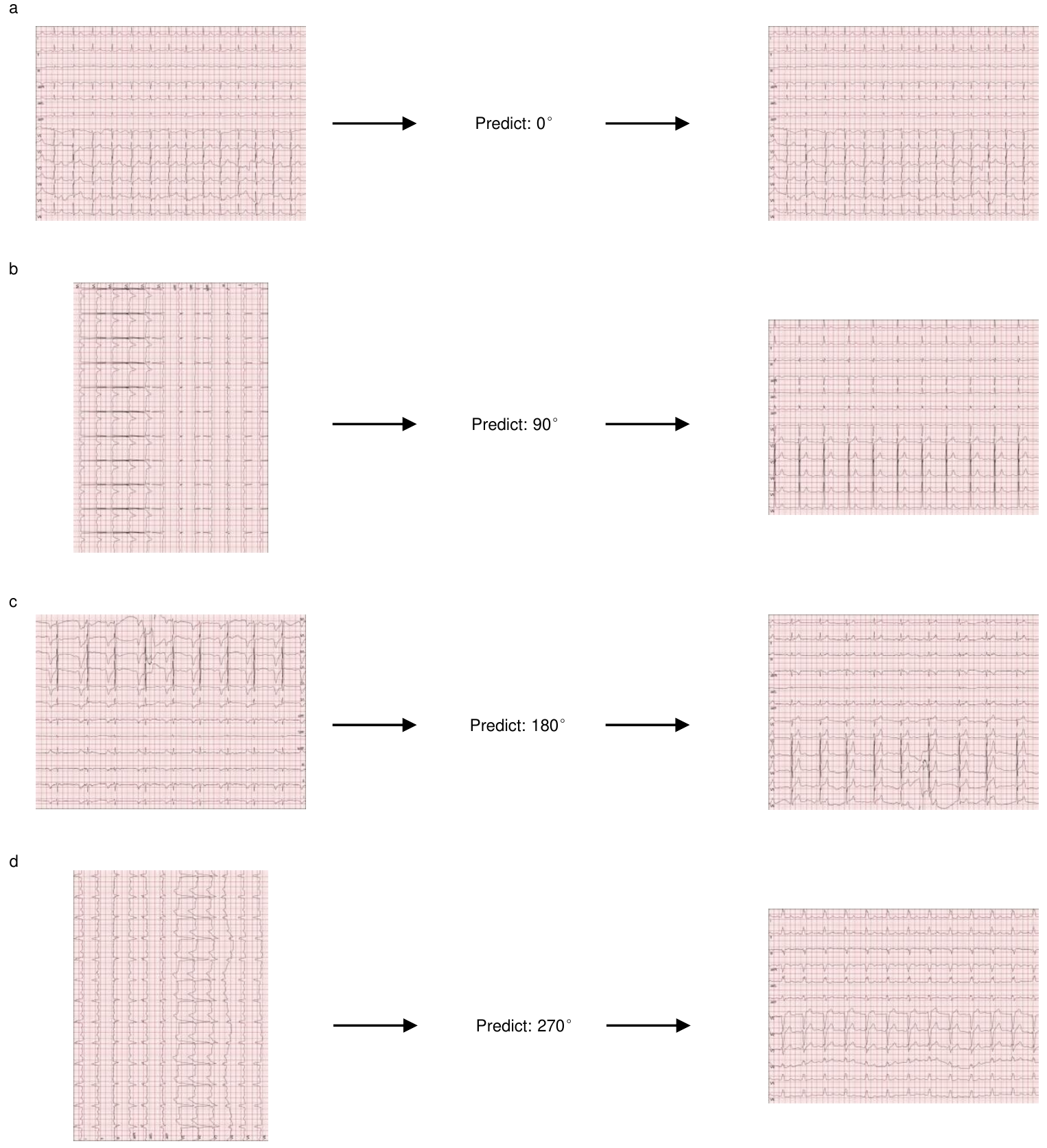}
    \caption{{\bfseries{Results visualization of angle recognition task.}} The original image is shown on the far left. The predict results of our model are shown in the middle. The corrected ECG image obtained after rotating the original image is displayed at the far right.}
    \label{figext20}
\end{figure}

\begin{figure}[htp]
    \centering
    \includegraphics[width=11cm]{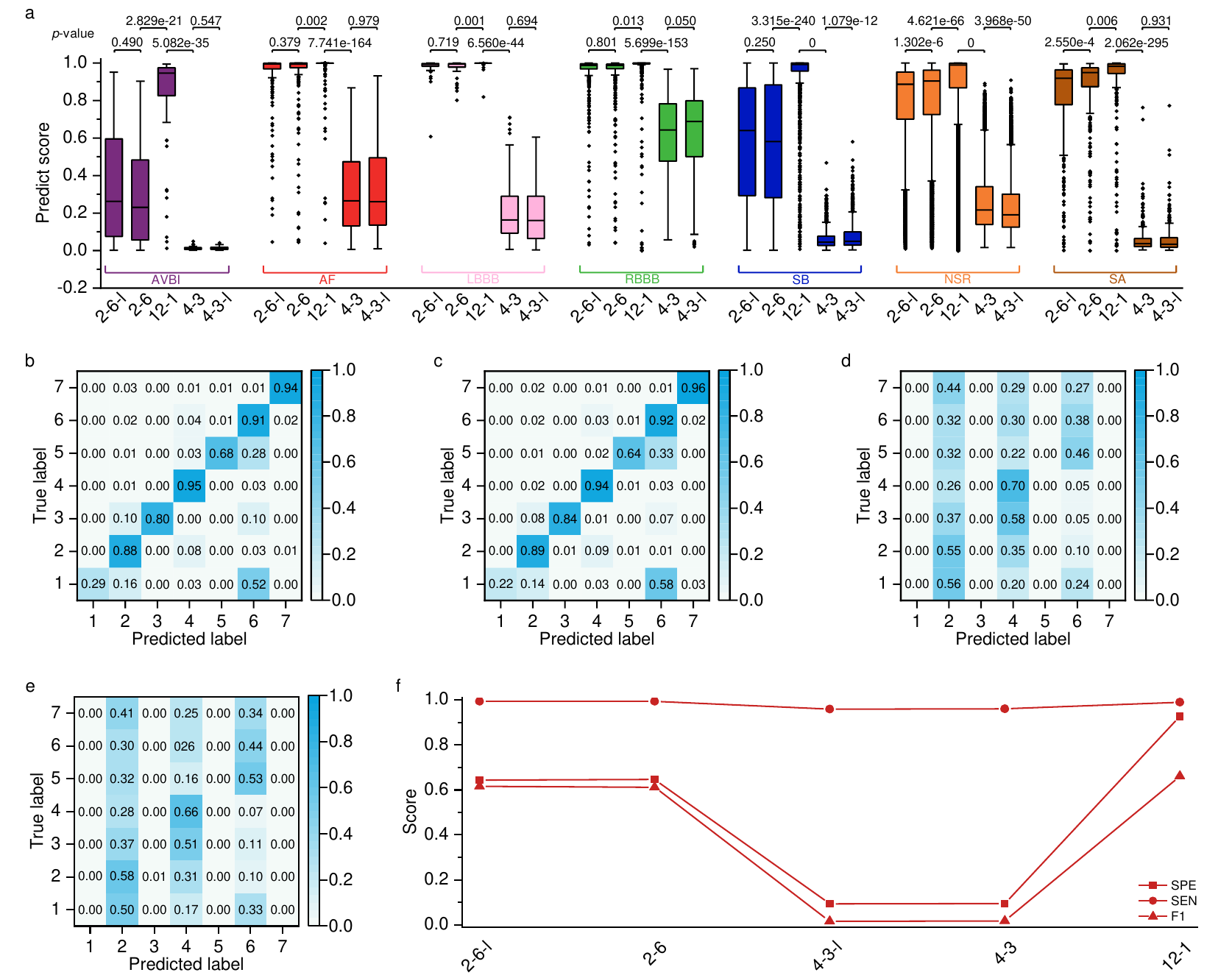}
    \caption{{\bfseries{Comparison results of ECG image at different lead positions.}} {\bfseries{a}}, boxplots of predict results from our model for different lead positions. 2-6-l, 12-lead ECG signals of 10s are arranged in two columns and six rows (Extended Data Fig. \ref{figext16}k). 2-6, 12-lead ECG signals of 5s are arranged in 2 columns and 6 rows (Extended Data Fig. \ref{figext16}j). 12-1, 12-lead ECG signals of 10s are arranged in 12 columns and 1 rows (Extended Data Fig. \ref{figext16}i). 4-3, 12-lead ECG signals of 2.5s are arranged in 4 columns and 3 rows (Extended Data Fig. \ref{figext16}l). 4-3-l, 12-lead ECG signals of 10s are arranged in 4 columns and 3 rows (Extended Data Fig. \ref{figext16}m). On each box, the center line represents the median, and the bottom and top edges of the box represent the 25th and 75th percentiles, respectively. The extension beyond the box represents 1.5 times the interquartile range. Use diamonds to represent outliers. {\bfseries{b-e}}, Confusion matrices of our models on 2-6-l, 2-6, 12-1, 4-3, and 4-3-l ECG images, respectively. {\bfseries{f}}, SPE, SEN, and F1 performance curves with different lead distributions.}
    \label{figext21}
\end{figure}

\begin{figure}[htp]
    \centering
    \includegraphics[width=11cm]{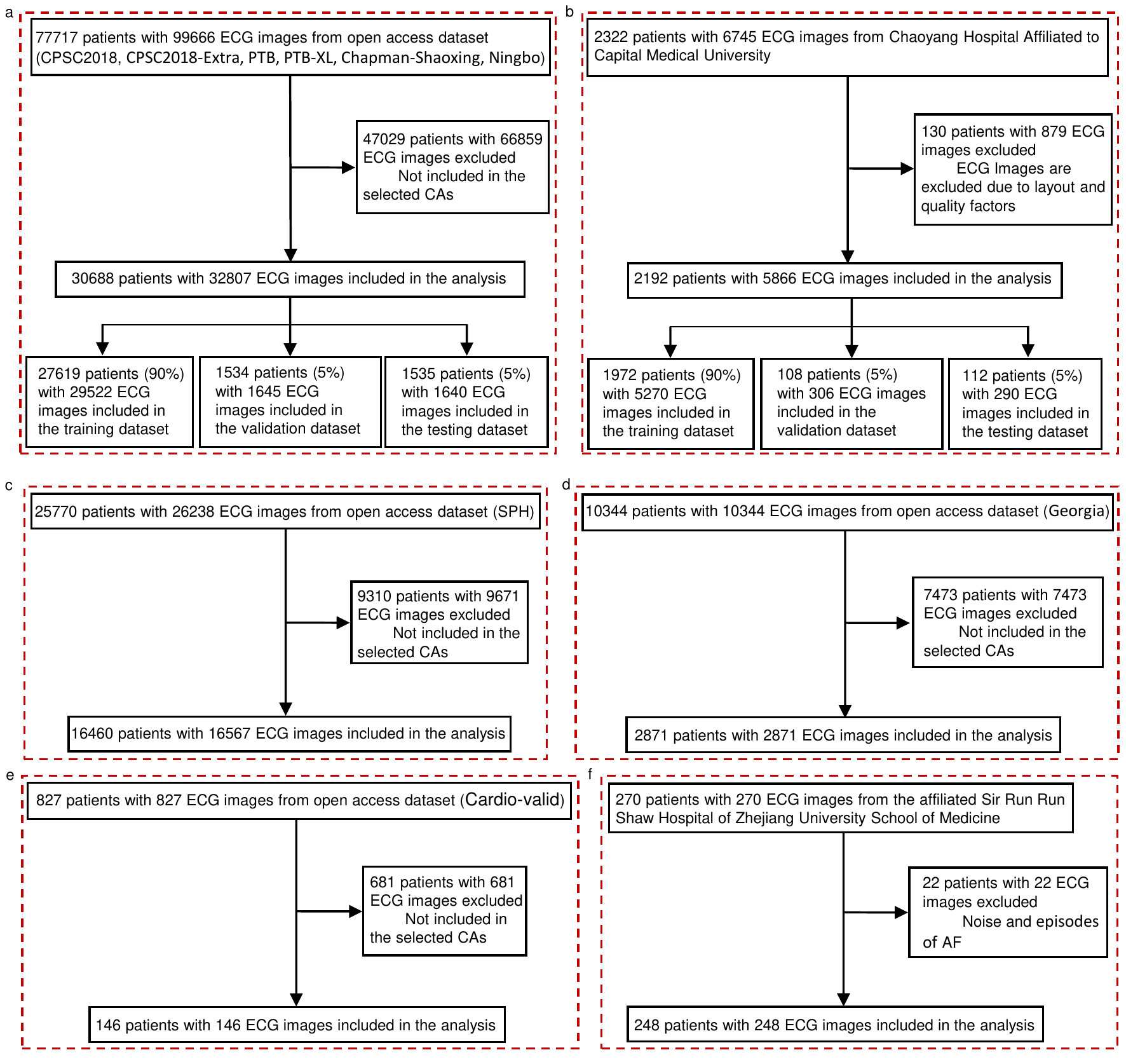}
    \caption{{\bfseries{Patient flow diagram of this study.}}}
    \label{figext22}
\end{figure}

In order to reduce the difficulty of real-world ECG image analysis and improve the accuracy of detection and screening tasks, we designed a reasonable decision process (Extended Data Fig. \ref{figext2}). First, we used an AI model \cite{tan2021efficientnetv2} to recognize the input image to determine whether it was an ECG image (Extended Data Fig. \ref{figext18}). The user was prompted to re-upload the image when the input image is not an ECG image (Extended Data Fig. \ref{figext1}a). Further manipulation was performed when the input image is an ECG image. Second, the AI model \cite{ge2021yolox} used for ECG image waveform region detection was constructed to obtain the waveform region of the uploaded ECG image and crop the detected waveform region(Extended Data Fig. \ref{figext19}). Third, the cropped waveform region was input to the angle recognition model \cite{tan2021efficientnetv2} to correct the angle of the waveform region to the horizontal (Extended Data Fig. \ref{figext20}). Fourth, the corrected waveform regions were analyzed by feeding into the AI model \cite{tan2021efficientnetv2} for ECG diagnosis and screening tasks. Fifth, each pixel contribution score in the ECG image was visualized based on the predicted results. Finally, the analysis results and pixel contribution visualization results were output for display. This solution effectively solves the challenges and difficulties of ECG image analysis in the real world. As the solution progresses, more critical areas were provided to the AI model. In addition, the lightweight neural network architecture used could realize the rapid analysis of ECG images.

In summary, we integrated various AI image processing methods into one framework in a reasonable way to achieve efficient and accurate analysis of ECG images in the real world. The results of ablation experiments (Fig. \ref{fig5}d,e) showed that the cross-study of various methods could provided more significant features for downstream tasks, reduce the complexity of model calculation in downstream tasks, and improve the prediction performance of the model.

{\bfseries{Training details}} In recent years, transfer learning \cite{raghu2019transfusion} had been the focus of research in AI. The essence of transfer learning was to find the similarity between existing knowledge and new knowledge and achieve the final goal through the transfer of this similarity. Transfer learning could overcome problems such as less annotated datasets, weak computing capability of the platform, specific task requirements, etc. To our knowledge, there was no framework that could be directly used for transfer in ECG image analysis tasks. At present, most researchers used the weight of the pre-trained model on ImageNet dataset \cite{deng2009imagenet} as the initial weight to realize transfer learning. Therefore, AI models that completed pre-trained in the ImageNet dataset were used for transfer learning in this study (Extended Data Fig. \ref{figext2}).

The visualization of a large number of open-source ECG signal datasets could provide strong data support for the construction of the ECG analysis task pretraining model. The AI model could learn the universal features of the ECG image by training the model with a large number of ECG images. In this way, the AI model transforms the general features extracted from the input images from natural images into ECG images. The transformation of features extracted makes the AI model overcome the influence brought by the image characteristics, which was conducive to the training of other tasks in ECG image analysis (such as segmentation, denoising, digitization, etc.). our model trained on the large paper-based ECG data was expected to be the transfer learning model architecture that can be directly used for the ECG image analysis task in the future.

{\bfseries{Data preprocessing and training settings}} Data augmentation technology played a crucial role in deep learning \cite{shorten2019survey}. Data augmentation could increase the diversity of samples and improved the generalization and robustness of the model. In addition, data augmentation could prevent the model from learning information irrelevant to the goal and reduce the risk of overfitting. Therefore, RandomHorizontalFlip\footnote[1]{\href{http://pytorch.org/vision/master/generated/torchvision.transforms.RandomHorizontalFlip.html}{http://pytorch.org/vision/master/generated/torchvision.transforms.RandomHorizontalFlip.html}}, RandomResizedCrop\footnote[2]{\href{http://pytorch.org/vision/main/generated/torchvision.transforms.RandomResizedCrop.html}{http://pytorch.org/vision/main/generated/torchvision.transforms.RandomResizedCrop.html}}, and Normalize\footnote[3]{\href{http://pytorch.org/vision/main/generated/torchvision.transforms.Normalize.html}{http://pytorch.org/vision/main/generated/torchvision.transforms.Normalize.html}} were used to perform data enhancement on the input image before training.

The model was optimized end-to-end based on Adam \cite{kingma2015adam}. The batch size is 32. The weight decays is set to 1e-4. The initial learning rate is set to 1e-3. Dynamic adjustment of the learning rate was realized by monitoring AUC on the validation set. The learning rate decreases by 0.3 times when the AUC did not rise within 5 epochs. The early stop was enabled to prevent overfitting. The training of the model was terminated when the learning rate was lower than 1e-6. At the same time, the model with the largest AUC on the validation dataset was saved as the final model. All models were built using the PyTorch (1.7) framework \cite{paszke2019pytorch} and trained on NVIDIA RTX 2080Ti 11G.

{\bfseries{Model and system deployment}} AI model and diagnostic strategy were deployed as cloud service via Django. Accessible Application Programming Interface (API) was provided to enable a mobile device to access cloud service. The information transmitted between the server and the mobile device was based on HTTP. An HTTP request included a title and a POST. The information on authorization was included in the title. Authorization was the only token provided by the server.

Furthermore, a simple user interface and flow path were designed so that users can quickly learn to use the system (Extended Data Fig. \ref{figext1}b). The flow path of use was designed as follows. First, an ECG image was obtained based on the user taking a paper-based ECG with a mobile device. Second, users uploaded the ECG image from mobile devices to the cloud service via the Internet. Third, the uploaded ECG images were automatically analyzed based on the AI model. Finally, the analysis results were sent to the mobile device via the Internet and showed the AI analysis details. In addition, the user could choose to send the analysis results to the doctor via the mobile device for review when the user had doubts about the analysis results. Users could get analysis results on mobile phones with Internet access on a global scale based on the above flow path.

{\bfseries{Evaluation methods}} Pearson correlation coefficients were used to assess the difference in diagnosis between our model and the cardiologist. Pearson correlation coefficient is calculated as:

\begin{equation}
P = {\dfrac{\sum_{i=1}^N (g_i - \bar{g})(y_i - \bar{y})}{\sqrt{\sum_{i=1}^N (g_i - \bar{g})^2}\sqrt{\sum_{i=1}^N (y_i - \bar{y})^2}}}
\end{equation}

\noindent Where $N$ is the number of samples, $g_i$ is the ground truth of cardiologist diagnosis for the $i$th sample, $\bar{g}$ is the average value of cardiologist diagnoses, $y_i$ is the predict of our model for the $i$th sample, $\bar{y}$ is the average value of predict of our model.

Accuracy (ACC), Sensitivity (SEN), Specificity (SPE), and F1-score (F1) were calculated to evaluate the performance of our model and system. In addition, ROC curves of AI models at different fine-tuning stages were plotted to estimate AUC. TP, FN, TN, and FP represent true positive, false negative, true negative, and false positive, respectively.

\begin{equation}
ACC = {\dfrac{TP+TN}{TP+FP+FN+TN}}
\end{equation}
\begin{equation}
SEN = {\dfrac{TP}{TP+FN}}
\end{equation}
\begin{equation}
SPE = {\dfrac{TN}{FP+TN}}
\end{equation}
\begin{equation}
F1 = {\dfrac{2*TP}{2*TP+FP+FN}}
\end{equation}

% Please refer to Journal-level guidance for any specific requirements.
% \backmatter
\section*{Acknowledgments}
This work was supported in part by the National Natural Science Foundation of China (No.62102008), the Hebei Science and Technology Project (22377785D), the Tianjin Municipal Natural Science Foundation (No. 21JCZDJC01080), the Tianjin Key Medical Discipline (Specialty) Construction Project (No. TJYXZDXK\-029A), and the Jiont Fund for Medical Artificial Intelligence (MAI2022Q011).

\subsection*{Data availability}
Data on arrhythmias used in this paper are publicly available and can be accessed as follows: CinC2021 Challenge \cite{goldberger2000physiobank,alday2020classification,reyna2021will} and SPH \cite{liu2022large}. Restrictions apply to the availability of the AF screening dataset (that is, Chaoyang dataset), which are thus not publicly available. Please email all requests for academic use of raw and processed data to hongshenda@pku.edu.cn. Requests will be evaluated based on institutional and departmental policies to determine whether the data requested is subject to intellectual property or patient privacy obligations. Data can only be shared for noncommercial academic purposes and will require a formal data use agreement. ECG images of arrhythmias used in this study will be published in the future. If you would like to use this data before publishing, please contact the corresponding author (hongshenda@pku.edu.cn).

\bibliography{reference}

\section*{Extend Data}

\begin{table}[htp]
	\caption{Characteristics of the datasets used in this study.}
	\centering
	\scalebox{0.6}{
		\begin{tabular}{cccccccc}
   		\hline
			Data scoure & Research type & Data type
 & Plot or not & No. of patients & No. of images & Task & No. of images per subject \\
			\hline
			CPSC2018 & Arrhythmia detection study & Signal & Yes & 4508 & 5723 & CAs detection & 1 (1)\\
			CPSC2018-extra & Arrhythmia detection study & Signal & Yes & 6 & 7 & CAs detection & 1 (0)\\
			PTB & Arrhythmia detection study & Signal & Yes & 85 & 988 & CAs detection & 12 (1)\\
			PTB-xl & Arrhythmia detection study & Signal & Yes & 7066 & 7066 & CAs detection & 1 (0)\\
			Georgia(External test cohort) & Arrhythmia detection study & Signal & Yes & 2871 & 2871 & CAs detection & 1 (0)\\
			Chapman-shaoxing & Arrhythmia detection study & Signal & Yes & 4915 & 4915 & CAs detection & 1 (0)\\
			Ningbo & Arrhythmia detection study & Signal & Yes & 14108 & 14108 & CAs detection & 1 (0)\\
			SPH (External test cohort) & Arrhythmia detection study & Signal & Yes & 16460 & 16567 & CAs detection & 1 (0)\\
			Cardio-valid (External test cohort) & Arrhythmia detection study & Signal & Yes & 146 & 146 & CAs detection & 1 (0)\\
			Chaoyang & Arrhythmia screening study & Image & No & 2192 & 5866 & CAs screening & 3 (2)\\
            Shaoyuifu (External test cohort) & Arrhythmia screening study & Image & No & 248 & 248 & CAs screening & 1 (0)\\
			\hline
	\end{tabular}}\label{extab1}
\end{table}

\begin{table}[htp]
	\caption{AI Model performance on the hold-out test dataset and external validation dataset. Numbers in parentheses represent the 95$\%$CI.}
	\centering
	\scalebox{0.65}{
	%% \tablesize{} %% You can specify the fontsize here, e.g., \tablesize{\footnotesize}. If commented out \small will be used.
	\begin{tabular}{ccccccc}\\
	\hline
	Dataset & Label & ACC & SEN & SPE & F1 & AUC\\
	\hline
	Held-out test & AVBI & 0.977(0.970-0.984) & 0.805(0.787-0.824) & 0.986(0.980-0.991) & 0.770(0.750-0.790) & 0.986(0.981-0.992)\\
	& AF                 & 0.998(0.996-1.000) & 1.000(1.000-1.000) & 0.998(0.996-1.000) & 0.988(0.983-0.993) & 1.000(0.999-1.000)\\
	& LBBB               & 0.998(0.995-1.000) & 0.870(0.854-0.885) & 0.999(0.998-1.000) & 0.909(0.896-0.923) & 0.998(0.995-1.000)\\
	& RBBB               & 0.988(0.983-0.993) & 0.960(0.950-0.969) & 0.991(0.986-0.995) & 0.935(0.923-0.946) & 0.997(0.994-1.000)\\
	& SB                 & 0.972(0.964-0.979) & 0.945(0.934-0.956) & 0.983(0.977-0.989) & 0.951(0.941-0.961) & 0.995(0.992-0.999)\\
	& NSR                & 0.960(0.951-0.970) & 0.951(0.941-0.961) & 0.967(0.959-0.976) & 0.955(0.945-0.964) & 0.991(0.987-0.996)\\
	& SA                 & 0.991(0.987-0.996) & 0.953(0.943-0.963) & 0.996(0.993-1.000) & 0.959(0.949-0.968) & 0.998(0.997-1.000)\\
	& Weighted mean & 0.992(0.988-0.996) & 0.889(0.875-0.904) & 0.995(0.992-0.998) & 0.902(0.888-0.916) & 0.996(0.993-0.999)\\
	\hline
	SPH & AVBI & 0.978(0.976-0.980) & 0.828(0.822-0.934) & 0.979(0.977-0.981) & 0.227(0.221-0.234) & 0.985(0.983-0.987)\\
	& AF       & 0.999(0.998-0.999) & 0.993(0.992-0.994) & 0.999(0.999-0.999) & 0.970(0.967-0.973) & 1.000(1.000-1.000)\\
	& LBBB     & 1.000(0.999-1.000) & 1.000(1.000-1.000) & 1.000(0.999-1.000) & 0.955(0.952-0.958) & 1.000(1.000-1.000)\\
	& RBBB     & 0.990(0.988-0.991) & 0.981(0.978-0.983) & 0.990(0.989-0.992) & 0.856(0.851-0.862) & 0.997(0.996-0.998)\\
	& SB       & 0.945(0.941-0.948) & 0.953(0.950-0.956) & 0.944(0.941-0.948) & 0.748(0.741-0.754) & 0.991(0.989-0.992)\\
	& NSR      & 0.931(0.927-0.935) & 0.921(0.917-0.925) & 0.983(0.982-0.985) & 0.957(0.954-0.960) & 0.989(0.988-0.991)\\
	& SA       & 0.991(0.989-0.992) & 0.920(0.916-0.924) & 0.993(0.991-0.994) & 0.832(0.826-0.938) & 0.995(0.994-0.996)\\
	& Weighted mean & 0.989(0.988-0.991) & 0.927(0.923-0.930) & 0.990(0.988-0.991) & 0.661(0.653-0.668) & 0.994(0.992-0.995)\\
	\hline
	Georgia & AVBI & 0.969(0.963-0.975) & 0.816(0.803-0.830) & 0.980(0.975-0.985) & 0.774(0.760-0.789) & 0.985(0.980-0.989)\\
	& AF           & 0.989(0.985-0.993) & 0.903(0.892-0.913) & 0.992(0.989-0.995) & 0.894(0.852-0.876) & 0.972(0.966-0.978)\\
	& LBBB         & 0.991(0.988-0.994) & 0.909(0.899-0.919) & 0.993(0.990-0.996) & 0.822(0.808-0.835) & 0.990(0.987-0.994)\\
	& RBBB         & 0.961(0.954-0.968) & 0.965(0.958-0.974) & 0.961(0.954-0.968) & 0.745(0.730-0.761) & 0.989(0.985-0.993)\\
	& SB           & 0.911(0.901-0.921) & 0.649(0.633-0.666) & 0.977(0.972-0.982) & 0.746(0.730-0.761) & 0.969(0.963-0.975)\\
	& NSR          & 0.879(0.868-0.891) & 0.921(0.912-0.931) & 0.815(0.801-0.829) & 0.902(0.892-0.913) & 0.948(0.940-0.956)\\
	& SA           & 0.975(0.970-0.981) & 0.833(0.820-0.846) & 0.989(0.986-0.993) & 0.858(0.846-0.871) & 0.994(0.991-0.996)\\
	& Weighted mean & 0.977(0.972-0.982) & 0.886(0.875-0.897) & 0.983(0.979-0.988) & 0.815(0.802-0.929) & 0.984(0.980-0.989)\\
	\hline
	Cardio-valid & AVBI & 0.932(0.892-0.971) & 0.679(0.606-0.751) & 0.992(0.977-1.000) & 0.792(0.728-0.855) & 0.992(0.979-1.000)\\
	& AF                & 0.932(0.892-0.971) & 0.923(0.882-0.965) & 0.932(0.893-0.971) & 0.706(0.635-0.777) & 0.982(0.961-1.000)\\
	& LBBB              & 0.945(0.910-0.981) & 0.900(0.853-0.947) & 0.957(0.925-0.989) & 0.871(0.819-0.923) & 0.977(0.954-1.000)\\
	& RBBB              & 0.890(0.842-0.939) & 0.735(0.667-0.804) & 0.938(0.900-0.975) & 0.758(0.691-0.824) & 0.933(0.894-0.972)\\
	& SB                & 0.966(0.937-0.994) & 0.938(0.900-0.975) & 0.969(0.942-0.996) & 0.857(0.803-0.912) & 0.984(0.965-1.000)\\
	& SA                & 0.945(0.910-0.981) & 0.784(0.720-0.848) & 1.000(0.000-1.000) & 0.879(0.828-0.930) & 0.991(0.976-1.000)\\
	& Weighted mean & 0.938(0.901-0.976) & 0.856(0.801-0.910) & 0.960(0.929-0.990) & 0.797(0.735-0.860) & 0.979(0.956-1.000)\\
	\hline
	\end{tabular}}\label{extab2}
\end{table}

\begin{table}[htp]
	\caption{Baseline}
	\centering
    \scalebox{0.7}{
	\begin{tabular}{cccccccc}
		\\
		\multicolumn{8}{l}{A. CPSC2018 dataset} \\
   		\hline
			Demographic & AVBI & AF & LBBB & RBBB & SB & NSR & SA\\
			\hline
			Age[years], mean(SD) & 66.8 (15.7) & 71.4 (12.5) & 69.7 (12.4) & 62.3 (17.2) & - & 41.6 (18.5) & -\\
			\hline
			Woman & 32\% & 43\% & 51\% & 35\% & - & 61\% & -\\
			\hline
      \\
		\multicolumn{8}{l}{B. CPSC2018-extra dataset} \\
   		\hline
			Demographic & AVBI & AF & LBBB & RBBB & SB & NSR & SA\\
			\hline
			Age[years], mean(SD) & 70.8 (9.8) & - & - & - & 72 (-) & - & 75 (14.1)\\
			\hline
			Woman & 75\% & - & - & - & 0\% & - & 50\% \\
			\hline
   \\
		\multicolumn{8}{l}{C. PTB dataset} \\
   		\hline
			Demographic & AVBI & AF & LBBB & RBBB & SB & NSR & SA\\
			\hline
			Age[years], mean(SD) & - & 53 (13.4) & - & - & - & 39.9 (13.6) & -\\
			\hline
			Woman & - & 0\% & - & - & - & 25\% & -\\
			\hline
   
   \\
		\multicolumn{8}{l}{D. PTB-xl dataset} \\
   		\hline
			Demographic & AVBI & AF & LBBB & RBBB & SB & NSR & SA\\
			\hline
			Age[years], mean(SD) & 62.3 (15.7) & 64.5 (16.3) & 69.6 (11.1) & - & 51.9 (16.7) & 51.5 (17.2) & 52.6 (17.9)\\
			\hline
			Woman & 33\% & 45\% & 43\% & - & 45\% & 55\% & 61\%\\
			\hline
   \\
		\multicolumn{8}{l}{E. Georgia dataset} \\
   		\hline
			Demographic & AVBI & AF & LBBB & RBBB & SB & NSR & SA\\
			\hline
			Age[years], mean(SD) & 69 (12.4) & 72.1 (10.6) & 68.8 (11.1) & 66.4 (15.5) & 60.3 (14.4) & 55 (14.9) & 52.4 (16.6)\\
			\hline
			Woman & 32\% & 29\% & 52\% & 32\% & 40\% & 54\% & 47\%\\
			\hline
   \\
		\multicolumn{8}{l}{E. Chapman-shaoxing dataset} \\
   		\hline
			Demographic & AVBI & AF & LBBB & RBBB & SB & NSR & SA\\
			\hline
			Age[years], mean(SD) & 67.3 (12.9) & 71.1 (11.2) & 72.6 (10.8) & 66.3 (15.6) & 55.8 (13.6) & 52.2 (15.9) & 47.7 (20.9)\\
			\hline
			Woman & 26\% & 34\% & 38\% & 28\% & 40\% & 57\% & 52\%\\
			\hline
   \\
		\multicolumn{8}{l}{F. Ningbo dataset} \\
   		\hline
			Demographic & AVBI & AF & LBBB & RBBB & SB & NSR & SA\\
			\hline
			Age[years], mean(SD) & 65.5 (14.3) & - & 69.8 (10.4) & 66.9 (13.5) & 55.7 (13.7) & 49.8 (18) & 39.1 (24.5)\\
			\hline
			Woman & 28\% & - & 57\% & 23\% & 41\% & 57\% & 52\%\\
			\hline
    \\
		\multicolumn{8}{l}{G. SPH dataset} \\
   		\hline
			Demographic & AVBI & AF & LBBB & RBBB & SB & NSR & SA\\
			\hline
			Age[years], mean(SD) & 65.2 (17.2) & 64.2 (12.3) & 67.8 (13.1) & 62.4 (15.4) & 53.2 (13.4) & 47.1 (14.0) & 43.6 (16.4)\\
			\hline
			Woman & 65\% & 64\% & 68\% & 62\% & 53\% & 47\% & 53\%\\
			\hline
   \\
        \multicolumn{4}{l}{H. Chaoyang dataset} & \multicolumn{4}{l}{I. Shaoyifu dataset} \\
   		\cline{1-3} \cline{5-7}
			Demographic & NAF & YAF &  & Demographic & NAF & YAF \\
			\cline{1-3} \cline{5-7}
			Age[years], mean(SD) & 61.9 (17.2) & 63.4 (10.7) &  & Age[years], mean(SD) & 32.4 (10.2) & 63.4 (11.0)\\
			\cline{1-3} \cline{5-7}
			Woman & 48\% & 43\% &  &Woman & 52\% & 47\%\\
			\cline{1-3} \cline{5-7}
	\end{tabular}}\label{extab3}
\end{table}

\begin{table}[htp]
	\caption{Table of terminology.}
	\centering
	\scalebox{1}{
		%% \tablesize{} %% You can specify the fontsize here, e.g., \tablesize{\footnotesize}. If commented out \small will be used.
	\begin{tabular}{cc}
	\hline
	abbreviation & full title\\
 	\hline
        CA & cardiac abnormalities\\
        CVD & cardiovascular disease\\
 	ECG & electrocardiogram\\
   	AI & artificial intelligence\\
        AUC & area under the receiver operating curve\\
        CAI & computer-assisted intervention\\
        ROC & receiver operating characteristic\\
        SEN & sensitivity\\
        SPE & specificity \\
        F1 & F1 score\\
        CI & confidence interval\\
        DCA &  decision curve analysis\\
        AVBI & first degree atrioventricular block\\
        AF & atrial fibrillation\\
        LBBB & left bundle branch block\\
        RBBB & right bundle branch block\\
        SB & sinus bradycardia\\
        NSR & sinus rhythm\\
        SA & sinus tachycardia\\
        NAF & sinus rhythm without AF\\
        YAF & sinus rhythm before AF\\
        IRBBB & incomplete right bundle branch block\\
        CRBBB & complete right bundle branch block\\
        Grad-CAM & gradient-weighted class activation map\\
        ECG image & photo, print screen, or scan from paper-based ECG\\
	\hline
	\end{tabular}}\label{extab}
\end{table}

\end{document}